\pdfoutput=1
\documentclass[11pt]{article}

\usepackage{acl}

\usepackage{times}
\usepackage{latexsym}
\usepackage{hyperref}
\usepackage{listings}
\usepackage{enumitem}
\usepackage{authblk}
\usepackage{tikz}
\definecolor{lightgray}{rgb}{0.93, 0.93, 0.93}
\tikzstyle{arrow} = [thick,->,>=stealth]
\definecolor{bluex}{RGB}{55,126,184}

\usepackage[T1]{fontenc}
\usepackage[utf8]{inputenc}

\usepackage{microtype}

\usepackage{inconsolata}
\usepackage{color,soul,colortbl}
\usepackage{xcolor}
\usepackage{booktabs}
\usepackage{amsmath}

\newcommand*\samethanks[1][\value{footnote}]{\footnotemark[#1]}
\makeatletter
\renewcommand\AB@affilsepx{ \protect\Affilfont}
\makeatother

\definecolor{cobalt}{rgb}{0.0, 0.28, 0.67}
\definecolor{purple}{rgb}{0.725, 0.36, 0.956}

\usepackage[obeyDraft]{todonotes}

\title{Exploration of Plan-Guided Summarization for Narrative Texts:\\ the Case of Small Language Models}

\author[1]{\bf Matt Grenander\thanks{Work done during an internship at AWS AI Labs.}\thanks{Equal contribution.}}
\author[2]{\bf Siddharth Varia\samethanks}
\author[3]{\bf Paula Czarnowska\thanks{Work done while at AWS AI Labs.}}
\author[2]{\authorcr \bf Yogarshi Vyas}
\author[2]{\bf Kishaloy Halder}
\author[2]{\bf Bonan Min}
\affil[1]{School of Informatics, University of Edinburgh \authorcr}
\affil[2]{AWS AI Labs} 
\affil[3]{PolyAI \authorcr}
\affil[ ]{\texttt{matt.grenander@ed.ac.uk, paula.czarnowska@poly-ai.com} \authorcr}
\affil[ ]{\texttt{\{siddhvar, yogarshi, kishaloh, bonanmin\}@amazon.com}}

\begin{document}
\maketitle
\begin{abstract}
Plan-guided summarization attempts to reduce hallucinations in small language models (SLMs) by grounding generated summaries to the source text, typically by targeting fine-grained details such as dates or named entities.
In this work, we investigate whether plan-based approaches in SLMs improve summarization in long document, narrative tasks.
Narrative texts' length and complexity often mean they are difficult to summarize faithfully.
We analyze existing plan-guided solutions targeting fine-grained details, and also propose our own higher-level, narrative-based plan formulation.
Our results show that neither approach significantly improves on a baseline without planning in either summary quality or faithfulness.
Human evaluation reveals that while plan-guided approaches are often well grounded to their plan, plans are equally likely to contain hallucinations compared to summaries. 
As a result, the plan-guided summaries are just as unfaithful as those from models without planning.
Our work serves as a cautionary tale to plan-guided approaches to summarization, especially for long, complex domains such as narrative texts.\footnote{Code available at: \url{https://github.com/amazon-science/plan-guided-summarization}}

% Existing plan-guided solutions mainly target fine-grained details within text and base their plans on QA pairs or entity chains. In this work we hypothesize that, while this setting works well for shorter texts, it falls short on challenging longer texts with a highly abstractive or narrative structure.
% To test this hypothesis, we propose an alternative plan formulation technique based on \emph{higher-level} planning. Our plans are based on \emph{sub-events} which focus on relevant occurrences without finer details such as dates and numbers.
% We show that our method generates summaries that are both more faithful and of higher quality than baselines in long-text settings.
% Using an NLI-based faithfulness metric, our best method outperforms comparable baselines with no planning by almost 10 points and baselines with only lower-level planning by 6 points on a challenging, short story summarization dataset.
% Our results are more mixed on a TV show transcript summarization task, and we highlight difficulties in assessing faithfulness in highly abstractive domains.
\end{abstract}

\section{Introduction}

% Paragraph 1: Problem
Modern summarization approaches based on language models generate increasingly fluent and useful summaries. 
However, both large language models (LLMs), and small language models (SLMs) of less than 4 billion parameters are prone to ``hallucinations'', where entities, dates, or assertions in predicted summaries do not faithfully reflect the source material \cite{kryscinski-etal-2019-neural, maynez-etal-2020-faithfulness, lin-etal-2022-truthfulqa, ziwei_etal_2023, wang2023survey}.
Such errors create trust and accountability issues, as users cannot rely on the outputs of the summarizer \cite{hci1}.

% Paragraph 2: Recent-works to address it and their shortcomings
\textit{Plan-guided summarization} is a leading approach for increasing faithfulness in SLM-based summarizers.
%The approach
It grounds summaries in a \emph{plan} which reflects relevant material from the source text. 
Models are fine-tuned to generate both plans and summaries, with the expectation that the plan is highly accurate and that the summary reflects planning content.
Since it involves fine-tuning, plan-guided summarization is typically applied to SLMs, where full fine-tuning is more feasible compared to LLMs.
Recent plan-guided summarization works have explored a diverse range of plan types; e.g., entity chains \cite{narayan-etal-2021-planning}, question-answer pairs \cite{conditional-narayan-etal-2023, pagnoni-etal-2023-socratic}, sentence fragments \cite{adams-etal-2023-generating}, semantic graphs \cite{hua-etal-2023-improving}, and salient noun phrases \cite{deutsch-roth-2023-incorporating}.

\begin{figure}[t!]
    \centering
    \small
    \begin{tabular}{|p{0.94\linewidth}|}
\hline
\vspace{0.1cm}
\ldots The humanoids rally around their leader, \textcolor{red}{who is wearing the same red sash and red headdress as the man.} The humanoid leader marches over to Linden and Split’s ship.\\ \\

Gravgak, the guard of the humanoid leader, confronts Linden and Split. He is worried that the sponge trees, which are camouflaging the warriors, will attack the humanoids again. He wants Linden and Split to repeat the “see-o-see-o” siren noises to keep the humanoids safe.\\ \\

\textcolor{red}{When Linden and Split repeat the siren, the warriors run back into their hiding places.} Tomboldo is the leader of the humanoids, and he is wearing the red sash and headdress. \textcolor{red}{Linden gives Tomboldo a gift, a musical medallion that plays a tune when it is touched.} \ldots
\vspace{0.1cm} \\
\hline
    \end{tabular}
    \caption{A predicted summary from Phi-3.5-mini \cite{phi3} after fine-tuning on SQuALITY \cite{wang-etal-2022-squality}. Red text marks hallucinated fragments.}
    \label{fig:motivating-example}
\end{figure}

% In this work, we investigate to what degree can planning reduce hallucination in summarizing long document, narrative-based text.
In this work, we investigate whether planning can reduce hallucination in summarizing long, narrative-based text with SLMs.
Narrative texts are often much longer than common summarization domains such as news \cite{wang-etal-2022-squality,chen-etal-2022-summscreen}, and requires a deep understanding of states, events, and temporal and causal relations \cite{kryscinski-etal-2022-booksum,kim2024fables}.
Many works have shown both SLMs and LLMs struggle at understanding long text \cite{levy-etal-2024-task, liu-etal-2024-lost}, including narrative-based texts such as novels \cite{kim2024fables}.
% Despite increasingly powerful performance from both SLMs and LLMs, many works have shown they struggle at understanding long text \cite{levy-etal-2024-task, liu-etal-2024-lost}, including in narrative-based texts such as novels \cite{kim2024fables}.

In Figure \ref{fig:motivating-example} we show an example output after fine-tuning Phi-3.5-mini \cite{phi3} on the SQuALITY dataset \cite{wang-etal-2022-squality}.
While it correctly generates many aspects, the summary also contains factual errors, highlighted in red. 
% It conflates the identity of characters (the man wearing the red sash and headdress is in fact the leader), generates non-existent events (Linden and Split do not repeat the siren),
It conflates the identity of characters (the leader and the man wearing the red sash and headdress are the same person), generates non-existent events (Linden and Split do not repeat the siren),
and inserts correct events at the wrong time (Linden presents Tomboldo a gift much earlier in the story).

These aspects may be challenging to address for existing planning-based approaches, which typically model fine-grained details such as specific entities \cite{narayan-etal-2021-planning, conditional-narayan-etal-2023}, sentence or subsentence-level details \cite{adams-etal-2023-generating, pagnoni-etal-2023-socratic}, or relationships between concepts and entities \cite{hua-etal-2023-improving}.
While establishing key characters, dates and quantities is crucial for a summary, they do not reflect the overall narrative structure of a text. 

% To that end, we train and evaluate several planning-based approaches, including QA-based planning methods \cite{narayan-etal-2023-conditional}, and our own proposed narrative planning method.
To that end, we propose a \textit{narrative planning} method based on \emph{sub-events} \cite{daniel-etal-2003-sub}, corresponding to high-level events that outline the narrative structure of a text. 
% Since manually annotating training plans is infeasible, we generate synthetic training data using \tocheck{Claude Sonnet 3.5 (v1)}\footnote{https://claude.ai/} (hereafter Sonnet 3.5) in a single pass, reserving training and inference to the SLMs}.
Since manually annotating training plans is infeasible, we use Claude Sonnet 3.5 (v1)\footnote{\url{https://claude.ai/}}(hereafter Sonnet 3.5) to generate synthetic training data in a single pass, leaving training and inference to the SLMs.
% Crucially, since LLMs are known to hallucinate details \cite{llm-hallucination}, we prompt the model to only extract the main events of the text, excluding the possibly hallucinated details.
We also experiment with QA-based planning methods \cite{narayan-etal-2023-conditional}, and mixing both narrative plans with QA-based ones.
% Lastly, we also experiment with mixing both narrative plans with QA-based ones.
%We then  
For each method, we fine-tune Transformer-based decoder models to generate both plans and gold summaries. 
An overview of our plan formulation is shown in Figure \ref{fig:viz-ex}.
% This is like distilling from a large LLM  \cite{hinton2014distilling}

% Main results
% In contrast to prior works on plan-based summarization, we find that planning-based methods neither from fine-grained nor narrative-based plans, scoring similarly to baselines on automatic metrics.
Contrary to earlier studies on plan-based summarization, our findings indicate that neither fine-grained nor narrative-based planning improves performance, with planning-based methods achieving scores comparable to baseline approaches.
Since automatic summarization metrics are known to be unreliable \cite{kryscinski-etal-2019-neural, kim2024fables}, we also conduct a human evaluation of our models' outputs.
We find that models, with or without planning, all hallucinate at comparable rates, resulting in similar summary quality.
Although summaries are often well-grounded to their associated plans, the plans themselves may contain hallucinations, which subsequently lead to unfaithful summaries.

Analyzing synthetic plans from Sonnet 3.5 reveals that they are highly faithful to the source text, though on occasion miss relevant details from the source. % document.
We also see the effect of replacing the models' predicted plans with the high quality plans from Sonnet 3.5 and find that coverage (+10\%) and faithfulness (+6\%) improve according to our manual evaluation.
% I want to say: some people just say use LLMs! But, they are big and expensive, and SLMs are far more deployable / practical in low-resource settings.
% I could also add that our work serves as a cautionary tale to analogous approaches for LLMs such as CoT, that there are limits to how much CoT can improve faithfulness
While one possible conclusion may be that only LLMs such as Sonnet 3.5 are reliable enough to be trusted as summarizers, SLMs are interesting in their own right as their smaller size allows them to be far more deployable and practical in low-resource settings.
For this reason, methods that promise higher faithfulness in SLMs, such as plan-guided summarization, are also appealing. However, despite encouraging results in prior work, our findings suggest that there are limitations in planning-based approaches for complex, long document domains such as narrative text.

\begin{figure}[t!]
    \centering
    \include{example-fig}
    \vspace{-1cm}
    \caption{A visualization of the steps in creating training plans. First, coarse plans are generated with an LLM. Then, named entities are extracted from the relevant source sentences, and a question generation model generates a question for each entity. Lastly, low-quality questions are filtered out and the QA pairs are attached to the coarse plans.
   The figure shows one QA pair per plan point; in practice, the number of pairs may vary.}
   \vspace{-0.3cm}
    \label{fig:viz-ex}
\end{figure}

\begin{figure*}[t]
    \centering
    \small
    \begin{tabular}{|p{0.95\linewidth}|}
         \hline
         \vspace{0.1cm}
\ldots Linden and Campbell think their ship is out of sight, and watch a ritual that the man is performing to the setting sun. The crowd of people continues to increase, and Linden notices that the landscape is moving: trees are shifting in the ground. He and Campbell stay in the ship and observe the various types of clothing and the ritual itself, as well as the moving trees which seemed to be moving to attack the people. They are indeed warriors starting an attack, and started swinging weapons. Linden tells Campbell to start the siren on their ship to scare away the attackers, and the first man they'd seen, presumably the leader, starts towards the ship. Once they are close enough, it is obvious that the humanoids don't have eyebrows or eye lashes. Captain Linden hands the leader a medallion that plays a song, as a token of friendship. Tomboldo, the leader, starts a round of introductions through a lot of gesturing. Linden hopes to learn about the Serpent River through the people to understand its cultural significance, and these people start to ask about the siren noises. The warriors attack again and panic ensues, pushing the humans to use weapons this time. Gravgak, the guard who had been escorting the humans, is knocked down. As Linden tries to tend to him, Gravgak knocks him out with his club. Linden is unconscious for a few weeks, and Vauna, Tomboldo's daughter, spends a lot of time by the Captian's side. \ldots
        \vspace{0.2cm} \\  
         \hline 
\ldots \\
5. Linden and Campbell observe a ritual and moving trees. \\
6. Tree-like warriors attack the humanoid people. \\
7. Linden uses ship's siren to scare away attackers. \\
8. Humans make first contact with the planet's inhabitants. \\
9. Warriors attack again, prompting humans to use weapons. \\
10. Gravgak knocks Linden unconscious. \\ 
11. Linden recovers, forms connection with Vauna. \\ 
\quad Q: How long is Linden unconscious? A: a few weeks \\
\ldots \\
         \hline

    \end{tabular}
    \caption{An example fragment of a reference summary from SQuALITY  (top), with a fragment of an annotated plan from Sonnet 3.5 (numbered sentences) and QA pairs. 
    The full example can be found in Appendix \ref{sec:examples}.}
    \label{fig:plan-example}
\end{figure*}

\section{Related Work}

Summarization with a planning step has taken many forms in prior work.
Our most direct inspiration is \citet{conditional-narayan-etal-2023}, which we refer to as \textit{Blueprint QA} throughout this paper.
Blueprint QA uses question-answer pairs as a form of grounding for the predicted summary, with filters to ensure high-quality QA pairs.
We describe their method more thoroughly in Section \ref{sec:method}.

\citet{pagnoni-etal-2023-socratic} also presents plan-based summarization method based on QA pairs. 
Source-text sentences are converted into question-answer pairs as a pre-training objective.
Sentence-level questions may be viewed as a form of higher-level planning; however, in this work we argue that narrative structure traverses sentence boundaries, and advocate for generating plans from the entire document instead of a single sentence.

\citet{deutsch-roth-2023-incorporating} train a model to mark salient NPs in the source document, then generate a summary conditioned on the augmented document.
\citet{adams-etal-2023-generating} generate content plans from extractive elementary discourse units, then re-write and re-rank the candidate summaries. 
Although their model uses abstractive re-writing, the plans are ultimately based on extractive fragments which are unlikely to represent higher-level plans.

Perhaps closest to our work, in terms of motivation, is \citet{hua-etal-2023-improving} who present a distinct method using abstract meaning representation (AMR) graphs as a source of grounding.
Their work explicitly captures high-level information, but is very involved, requiring an AMR parser, a coreference resolution model and an additional module to align concepts and words, on top of the summarization model.
Our approach presents a simpler, less involved approach to high-level planning where narrative structure is represented as short sentences instead of complex AMR graphs.

Planning is used outside of summarization, and most relevant to our work is that of \citet{godbole-etal-2024-analysis}. 
They use planning to generate both paragraph descriptions and QA pairs in order to generate biographies about individuals.
However, their work focuses on using planning with retrieval to augment the model's parametric knowledge, while in our task, we are interested in knowing whether the model can understand and reason with long contexts in a narrative structure.
Similarly, \citet{shao-etal-2024-assisting} prompts LLMs to write Wikipedia articles by generating outlines, diverse perspectives and conversational QA-pairs as a plan.
Their work similarly focuses on using LLMs to retrieve helpful documents for writing the article, rather than summarizing long narrative text.

Lastly, \citet{chawla-etal-2024-investigating} also investigate planning techniques for generating knowledge-grounded dialogues, finding that content planning offers mixed results in improving dialogue quality.

% \citet{puduppully-etal-2019-data, data-to-text-puduppully-etal-2019} and \citet{pudupully-etal-2021-macro} study content selection and planning in summarizing tabular data paired with descriptive documents.
% However, this setting is very different from ours since we focus on long document summarization with narrative structure instead of tabular data.

\section{Method}\label{sec:method}
%In this section, we describe several planning-based approaches for summarization.
% Since summarization datasets do not come with plans, these are generated using specialized models or LLMs.
% We first detail in \ref{sec:gen-plans} how to create plans from summarization data including \citet{conditional-narayan-etal-2023}'s Blueprint QA approach and our own narrative-based planning formulation.
% We then explain in \ref{sec:training} how we fine-tune Transformer-based models to generate both plans and gold summaries.

In this work we experiment with an existing QA plan-based summarization method of \citet{conditional-narayan-etal-2023} (our re-implementation) and our novel \emph{narrative-based} planning methods that incorporate \textbf{coarse plans}. In this section we discuss the implementation details for each approach. Since summarization datasets do not come with plans, we generate them using specialized models or LLMs.
We first detail in \ref{sec:gen-plans} how to create plans from summarization data. % including \citet{conditional-narayan-etal-2023}'s Blueprint QA approach and our own narrative-based planning formulation.
We then explain in \ref{sec:training} how we fine-tune SLMs to generate both plans and gold summaries.

% As in prior work on plan-based summarization \cite{narayan-etal-2021-planning, hua-etal-2023-improving, conditional-narayan-etal-2023} we fine-tune Transformer-based models to generate both plans and gold summaries. 
% In this section, we first describe how we generate the training plans using LLMs and other specialized models (Section~\ref{sec:gen-plans}). 
% Next, we explain how we train summarization models using these plans (Section~\ref{sec:training}).

\subsection{Training Plans}\label{sec:gen-plans}
\paragraph{Blueprint QA} In \citet{conditional-narayan-etal-2023}'s Blueprint QA plan formulation, a question generation model overgenerates a sequence of questions relating to entities, times, and places in the document of interest, and the questions are then filtered for quality. 
We use SpaCy \cite{spacy} to extract named entities including dates, time expressions and quantities.
We then generate questions using MixQG\footnote{We use the T5-3B variant.} \cite{murakhovska-etal-2022-mixqg}, a state-of-the-art question generation model, with the extracted entity along with its source sentence and two preceding sentences as the context.
Lastly, we filter low quality QA pairs using Round-trip Consistency, Rheme,\footnote{We notice in certain cases that the rheme filters removes all QA pairs from the plan. In these cases, we do not apply the rheme filter.} and Coverage filters.

% Given a document $d$ and a summary $s$, we first obtain a high-level, coarse training plan $p$, consisting of plan points $p_1,\dots,p_m$. For each plan point, we create several question-answer pairs corresponding to relevant missing details.
% The number of QA pairs per plan point may vary and even be zero.

\begin{figure*}[t!]
    \centering
    \begin{tikzpicture}
\node [] (a) at (-1,0) {(a)};
\node [draw,
inner sep=10pt,
outer sep=0pt,
rounded corners=0.1cm,
fill=bluex!30,
right=0.5cm of a] (e2emodel)  {\small Model};

\node [draw,
inner sep=5pt,
align=left,
rounded corners=0.1cm,
right=0.5cm of e2emodel] (e2eplansumm) {\small <plan>\dots<\textbackslash plan>\\\small <summary>\dots<\textbackslash summary>};

\draw [arrow] (e2emodel) -- (e2eplansumm);

\node [right=0.5cm of e2eplansumm] (b) {(b)};
\node [draw,
inner sep=10pt,
outer sep=0pt,
rounded corners=0.1cm,
fill=bluex!30,
right=0.5cm of b] (multimodel) {\small Model};

\node [draw,
inner sep=5pt,
rounded corners=0.1cm,
above right = 0.1cm and 1.0cm of multimodel] (multiplan) {\small <plan>\dots<\textbackslash plan>};

\node [draw,
inner sep=5pt,
rounded corners=0.1cm,
below right = 0.1cm and 1.0cm of multimodel] (multisum) {\small <summary>\dots<\textbackslash summary>};

\draw [arrow] (multimodel) -- (multiplan.west);
\draw [arrow] (multimodel) -- (multisum.west);

\end{tikzpicture}
    \caption{The two training methods we explore: (a) In the End-to-End (E2E) setting, the model generates both plans and summaries in a single decoder pass. (b) In the Multi-Task setting, the model separately learns to generate plans and summaries. At inference time, we pre-fill the decoder with the <summary> token to prompt the model to generate the summary.}
    \label{fig:e2e-multitask-ex}
\end{figure*}

\paragraph{Coarse Plans} Since manually annotating sub-events is infeasible, we automatically generate the coarse plans containing these sub-events by prompting Sonnet 3.5, which is known for its strong text understanding and generation capabilities \cite{claude3}.
We prompt the model to extract key sub-events from the text, without including specific details which are more likely to be hallucinated \cite{ziwei_etal_2023, wang2023survey}.\footnote{See 
Appendix~\ref{sec:coarse-prompt} for the exact prompt.} Note that we only prompt Sonnet 3.5 in a one-shot basis to generate coarse plans, and we do not employ it during training or inference.

Each plan consists of numbered \emph{plan points}, where each plan point corresponds to one sub-event in the text.
Our definition of a sub-event is primarily inspired from \citet{daniel-etal-2003-sub}, but is also similar to the \emph{atomic claims} defined in \citet{gunjal-durrett-2024-molecular}.
Each sub-event should progressively and succinctly describe key events from the story, with minimal details such as dates or locations.
To encourage the LLM to output the right plan format, we provide in-context examples of the type of plan we expect~\cite{gpt3}.
Examples of the types of plans we expect can be seen in the in-context examples in Appendix~\ref{sec:coarse-prompt}.

We generate plans using temperature set to $1.0$, nucleus sampling disabled ($top\_p=1.0$), and limiting the output to 400 tokens.
These settings encourage diverse plans and consistent lengths.
Importantly, we prompt the model to construct the plan based on a gold summary $s$, instead of the source document $d$. 
This ensures the predicted plan matches the narrative of the gold summary.
An example is shown in Figure \ref{fig:plan-example} (without QA pairs).\footnote{The full coarse plan is shown in Appendix \ref{sec:examples}.}

\paragraph{Coarse Plans + QA}
% We expect coarse plans alone to be an insufficient plan schema since they are devoid of summary-worthy, lower-level details.
% We therefore augment each plan point with relevant fine-grained details in the form of QA pairs using a question generation model.
We also experiment with a setting which combines the benefits of both the higher-level, event-focused planning and the lower-level, detail-focused planning. To this end, we combine coarse and fine-grained QA into a single plan.
Question generation models typically require an answer and a context to generate an associated question. 
In our case, we do not directly input plan points to the question generation model, since plan points are designed to be free of the desired fine-grained details. % which we want to capture in QA pairs.
Therefore, our first step is to link plan points back to the original summary sentences, which contain more details suitable for fine-grained question generation.
The original source sentence that gave rise to the plan point contains relevant details abstracted away from the plan. 

To facilitate this linking, we ask Sonnet 3.5 to provide a citation for each plan point, similar to \citet{fierro-etal-2024-learning}. 
We number all sentences in the gold summary and ask the model to end each plan point with the relevant sentence number(s).\footnote{The exact prompt is shown in Appendix \ref{sec:coarse-prompt}.}
% We experiment with modifying the coarse planning prompt, by directing the LLM to cite sentences in the source text for each generated plan point, similar to \citet{fierro-etal-2024-learning}. % although our coarse plan formulation is based around narrative-based sub-events instead of QA pairs.
% We allow the LLM to match each plan point to multiple source text sentences as needed.

Given the linked source sentences, we apply the Blueprint QA procedure as before, using the linked sentences and two preceding sentences as context for the QG model.
After applying the Blueprint QA filters, each generated QA pair is then attached to the original plan point.
We then additionally filter out identical QA pairs \textit{across} plan points.
% An example of the plan is shown in Figure \ref{fig:plan-example}, with more examples in Appendix \ref{sec:examples}.

% We start by removing duplicates.
% For question-answer pairs that are exact matches, we keep the earliest mention of the QA pair and remove any subsequent ones.
% We also remove QA pairs with identical question text and different answers unless one answer is a substring of the other, in which case we keep the earlier QA pair, substituting in the longer answer if necessary.

\subsection{Training}\label{sec:training}
We use two training methods to generate plans and summaries, following \citet{conditional-narayan-etal-2023}.

\paragraph{End-to-End} In this setting, the model produces both plans and summaries in a single pass.
After processing document $d$, the model first produces the plan $p$, effectively modeling $\Pr(p \vert d)$.
The decoder then generates the summary $s$ based on both $p$ and $d$, modeling $\Pr(s \vert d, p)$.%in essence modeling $\Pr(s \vert d, p)$.
We prefix the plan with the string <plan>, and similarly prefix the summary with <summary>.
This allows the model to directly refer to both the plan and the source document via its attention mechanism when generating the summary. However, it comes at the cost of generating longer sequences.

\paragraph{Multi-Task} The end-to-end model may potentially struggle with extremely long generation when predicting both plan and summary in a single output.
For this reason, we also experiment with training the model to generate plans and summaries separately in a multi-task setup.
% In this setting, we investigate whether learning the narrative structure in one task can help with the main summarization task.
Each document is separately paired with the generated training plan and the gold summary, and the model learns to generate the plan and summary in separate tasks.
Although the model is less likely to be grounded in the summary in this setting, avoiding long generation may produce higher quality summaries.

\section{Experiments}
\subsection{Datasets}\label{sec:datasets}
We choose two long document summarization tasks with a clear event-focused narrative structure covering short stories and TV show transcripts.
% We expect our method to be most useful in long-document settings where planning can improve summary quality and reduce faithfulness errors.
% Thus, we focus on two long document summarization tasks with a clear event-focused narrative structure that suit our proposed high-level planning.

\paragraph{SQuALITY} is a long document summarization dataset covering short stories from Project Gutenburg~\cite{wang-etal-2022-squality}. 
The dataset contains 50/25/52 documents across training, validation, and test splits. Each document is paired with 4 summaries written by highly-qualified writers. 
On average, the dataset contains 7648 tokens per document and 591 tokens per summary.

\paragraph{SummScreen} is a long document summarization task based on TV episode transcripts \cite{chen-etal-2022-summscreen}. 
Summaries in this dataset tend to be very abstractive, as plot details are usually indirectly expressed by characters in dialogue. 
The document length also presents a challenge, as the dataset contains on average 7605 tokens per document and 114 tokens per summary.
Following prior work \cite{hua-etal-2023-improving, conditional-narayan-etal-2023}, we use the FullDreaming (FD) subset, which contains 3673/338/337 examples for training, development, and test sets across 88 TV shows.

\subsection{Model}
\paragraph{Base architecture} We use the Phi-3.5-mini model in all experiments \cite{phi3}, a 3.8B parameter decoder-only Transformer architecture pre-trained on 3.3T tokens.\footnote{We also experimented with LongT5 \cite{guo-etal-2022-longt5}, but got very poor summarisation results across all datasets, which we omit from this paper for reasons of space.} Despite its compact size, Phi-3.5-mini performs comparably to other leading LLMs in a variety of reasoning tasks \cite{phi3}. %while being much smaller and faster. %, making it a suitable choice for our experiments.
It also supports a 128K context window, allowing us to directly input long documents for the summarization task, making it an ideal candidate for an exploration of long document summarisation with smaller LLMs.
We include additional training details in Appendix \ref{sec:training-details}.

\subsection{Evaluation}
\label{subsec:eval}

% To evaluate the quality of the generated summaries we use 3 metrics: ROUGE \cite{lin-2004-rouge}, AlignScore \cite{zha-etal-2023-alignscore} and QAFactEval \cite{fabbri-etal-2022-qafacteval}.

\paragraph{{ROUGE}} is a standard set of summary quality metrics which measure word-level overlap with a reference summary \cite{lin-2004-rouge}. 
For SQuALITY, which includes 4 gold summaries per text, we compute ROUGE for each label and take the maximum of the four scores.
We report ROUGE-1, -2 and -L with Porter stemming.\footnote{We use the Hugging Face implementation from \url{https://huggingface.co/spaces/evaluate-metric/rouge}.}

\paragraph{AlignScore} is an NLI-based metric for measuring factual consistency between predicted summaries and their source text~\cite{zha-etal-2023-alignscore}. % reference summaries~\cite{zha-etal-2023-alignscore}. 
It splits the source document into 350 token chunks and then predicts whether each sentence in the predicted summary is entailed by any chunk in the source.
The final score is computed by taking the average value of the highest entailment scores for each sentence in the predicted summary.

% Since AlignScore relies on NLI scores between only one sentence and only one reference chunk, it is less suitable for evaluation of more abstractive summaries in which one summary sentence can capture details spread across multiple reference chunks. In such scenarios, the entailment scores for valid summary sentences can be misleadingly low, since none of the chunks in isolation entails them. Further, by-design, the metric is inherently biased towards 1) shorter, less composite sentences and 2) more verbose summaries that contain many sentences without much detail. This is because shorter sentences are more likely to directly map to only one chunk, while the many easy sentences can elevate the final score, which is a cross-sentence average. 
% We empirically observe such behavior in our results on SummScreen.
% \tocheck{Should we still include this observation?}

\paragraph{QAFactEval} is a QA-based metric measuring faithfulness between predicted summaries and source texts \cite{fabbri-etal-2022-qafacteval}.
It generates questions from the predicted summary, then measures the overlap between the answer using the input document vs. the predicted summary. 
Answer overlap is evaluated on a 1--5 scale, where higher is better.

\paragraph{FineSurE} is an LLM-based metric measuring faithfulness and summary quality \cite{song-etal-2024-finesure}.
This metric extracts and then aligns key facts from the generated and gold summaries, with the final score computed based on how many key facts are predicted as supported. We use Claude Sonnet 3.5 (v1) as the LLM.\footnote{Although using the same LLM for generating plans and evaluation is non-ideal as prior work has shown they are biased towards their outputs \cite{kim2024fables}, we are restricted in our use of available LLMs.}

\paragraph{Human Evaluation}
Automatic evaluation metrics are known to be unreliable \cite{kryscinski-etal-2019-neural, kim2024fables}, with different metrics capturing different dimensions of the outputs.
% Therefore, we also conduct human evaluation of selected summaries.
% We select 5 documents from SQuALITY and annotate all model settings.
% The annotations cover 8 models $\times$ 5 summaries $\times$ 4 evaluation metrics, for a total of 155 annotations (some metrics do not apply universally).
Therefore, we also conduct human evaluation of 5 randomly selected documents from SQuALITY. The annotation is conducted by 2 authors of this work, following several discussions on best practices and establishing an evaluation rubric.\footnote{Additional details can be found in Appendix \ref{sec:human-eval-app}.} It covers all settings: 8 models $\times$ 5 summaries $\times$ 4 evaluation metrics, for a total of 155 annotations (some metrics do not apply universally).

After reading each document, we list important high-level events from the story. We measure \textbf{coverage} as the proportion of high-level events from the story that are present in the generated summary.

For each predicted summary, we extract \textit{atomic claims} \cite{kim2024fables}, as we find using a finer level of granularity allows for more consistent annotation.
We aim to extract atomic claims that are indivisible minimal facts, which are context-independent and describe a property of an entity or a relationship between two entities.\footnote{Examples can be found in Appendix \ref{sec:human-eval-app}.} 
We measure \textbf{faithfulness} as the proportion of the model's atomic claims that are supported by the text, and \textbf{conciseness} as the proportion of atomic claims that are present in at least one reference summaries.

We are also interested is how well the predicted summary follows the generated plan.
For all settings generating plans, we compute \textbf{grounding} as the proportion of plan points that are present in the generated summary.

% Human annotation is conducted by authors of this work, following several discussions on best practices and establishing an evaluation rubric.\footnote{Additional details can be found in Appendix \ref{sec:human-eval-app}.}

% \paragraph{PRISMA} is an LLM-based metric measuring faithfulness and summary quality \cite{mahon-lapata-2024-modular}.
% It breaks down the gold and reference summaries into atomic facts, then queries the LLM whether each atomic fact is attested in the corresponding summary.
% These scores lead to `fact-precision' and `fact-recall' scores and used to compute an overall F1 score.
% PRISMA only assesses facts appearing in gold summaries as opposed to the source document, justifying the design choice by noting facts in the summary are likely to be the most relevant of the overall text.
% In our experiments, we follow \cite{mahon-lapata-2024-modular} and use GPT4 as the LLM; this choice also avoids evaluating with the same LLM as the one used to generate coarse plans.

\subsection{Compared Systems}
\paragraph{Phi-3.5-mini}
Our primary baseline is the Phi-3.5-mini model trained for summarization without using any plans.
% We train Phi-3.5-mini in a straightforward sequence-to-sequence setup, fine-tuning the model with the document as input and summary as target.
We fine-tune the model with the document and summary as input, with loss computed over the summary tokens.
We prefix each example with a basic summarization prompt.\footnote{The prompt can be found in Appendix \ref{sec:phi-baseline-prompt}.}

% \paragraph{Blueprint QA} We also compare to \citet{conditional-narayan-etal-2023} whose method is closely related to ours, albeit focusing on lower-level details.
% Since their implementation is not publicly available, we re-implement their E2E and multi-task settings using the Phi-3.5-mini model.
% We generate question-answer pairs using MixQG, then apply filtering based on round-trip consistency \cite{alberti-etal-2019-synthetic}, rheme, and coverage.

% \paragraph{Coarse Plans} In addition to training the model variant with both coarse and low-level details (Coarse Plans + QA), we also train model variants on high-level plans only.

% \paragraph{Coarse Plans + QA} This planning-style incorporates both coarse sub-event planning and fine-grained question answer pairs, as detailed in Section \ref{sec:method}.

\paragraph{Socratic} For SQuALITY we also report results from \citet{pagnoni-etal-2023-socratic} using the BART-large model \cite{lewis-etal-2020-bart}.
The model is pretrained with their \textbf{Socratic} objective on the Books3 corpus, from the Pile \cite{gao2020pile} before being fine-tuned on SQuALITY. 
% The Socratic objective involves generation of questions about selected masked sentences from the source text, followed by the generation of pseudo-summaries, conditioned on both the text and the generated questions.

\paragraph{Claude Sonnet 3.5 (v1)} 
We also report results obtained by directly prompting Claude Sonnet 3.5 (v1). 
We use a temperature of 0, max generation length of 512, and $top\_p=1.0$.
The full prompt is shared in Appendix \ref{sec:claude-baseline-prompt}.\footnote{The final prompt we use for Claude summarization does not ask the model to generate plans. We tried prompt variants that asked the model to plan-and-summarize, but these worked worse than our final prompt.}  Since we do not have access to the information about Claude training data, we cannot be sure that the model has not seen the documents in SummScreen or SQuALITY. For this reason, we add Claude results mostly for reference, rather than as a comparative baseline.

% RESULTS WITH FINESURE
\begin{table*}[t]
\small
\centering
\begin{tabular}{lr r rrrr}
\toprule											
Model	&	R-1	&	R-2	&	R-L	&	AlignScore	&	QAFactEval	&	FineSurE	\\
\midrule													
\multicolumn{7}{c}{\textsc{SQuALITY}}	\\				
\midrule													
Claude 3.5 Sonnet	&	47.52	&	12.07	&	20.03	&	\textbf{59.76}	&	1.42	&	\textbf{91.8}	\\
Phi-3.5-mini 	&	51.47	&	16.44	&	23.36	&	53.22	&	1.83	&	66.3\vspace{0.15cm} \\
Blueprint QA, E2E 	&	40.82	&	13.14	&	20.23	&	57.30	&	1.75	&	63.4	\\
Blueprint QA, Multi-Task 	&	51.54	&	16.35	&	23.75	&	53.85	&	1.76	&	67.4	\\
Socratic \cite{pagnoni-etal-2023-socratic}	&	46.31	&	14.80	&	22.76	&	-	&	-	&	-\vspace{0.15cm} \\
Coarse Plans, E2E	&	39.10	&	12.04	&	20.15	&	52.38	&	1.63	&	61.3	\\
Coarse Plans, Multi-Task	&	\textbf{51.66}	&	\textbf{16.58}	&	\textbf{24.27}	&	52.25	&	1.78	&	66.1\vspace{0.15cm} \\
Coarse Plans + QA, E2E	&	37.38	&	11.41	&	19.86	&	51.86	&	1.96	&	57.8	\\
Coarse Plans + QA, Multi-Task	&	50.57	&	16.12	&	23.90	&	54.39	&	\textbf{2.21}	&	67.1\vspace{0.15cm}\\													
Pre-filled Claude Plans, E2E	&	48.57	&	16.26	&	25.51	&	50.04	&	1.97	&	71.9	\\
\midrule													
\multicolumn{6}{c}{\textsc{SummScreen}}	\\												
\midrule													
Claude 3.5 Sonnet	&	28.87	&	7.27	&	14.91	&	\textbf{57.55}	&	1.69	&	\textbf{95.5}	\\
Phi-3.5-mini 	&	31.50	&	7.40	&	18.82	&	45.27	&	1.82	&	44.0\vspace{0.15cm} \\
Blueprint QA, E2E 	&	26.62	&	5.89	&	16.93	&	49.15	&	\textbf{1.88}	&	55.8	\\
Blueprint QA, Multi-Task 	&	31.45	&	6.92	&	18.54	&	46.23	&	1.77	&	45.7\vspace{0.15cm} \\
Coarse Plans, E2E	&	29.34	&	6.01	&	17.55	&	46.49	&	1.66	&	44.3	\\
Coarse Plans, Multi-Task	&	\textbf{31.98}	&	\textbf{7.55}	&	\textbf{19.02}	&	47.02	&	1.84	&	47.0\vspace{0.15cm}\\
Coarse Plans + QA, E2E	&	28.83	&	5.91	&	17.19	&	46.16	&	1.66	&	40.3	\\
Coarse Plans + QA, Multi-Task	&	31.58	&	7.30	&	18.62	&	45.93	&	1.85	&	46.2	\\
\bottomrule												
\end{tabular}
\caption{Results on SQuALITY and SummScreen.
Blueprint QA, E2E and Multi-Task, are our implementations of \citet{conditional-narayan-etal-2023}. Scores for Socratic \cite{pagnoni-etal-2023-socratic} are as  reported in their paper; all other scores are from our own implementations. }
% Coarse Plans is our proposed higher-level planning method, while + QA refers to adding fine-grained QA plans. 
%\textbf{Bold} indicates highest scores for each metric.}
\label{table:results}
\end{table*}

\section{Results}
Results on SQuALITY and SummScreen are shown in Table \ref{table:results}.

\paragraph{Summary Quality} In the multi-task setting, summary quality is generally comparable to the baseline method without planning.
The Coarse Plans, Multi-Task model achieves the highest ROUGE scores overall on both SQuALITY and SummScreen, though improvements over the Phi-3.5 baseline are all under 1 F1 point.
% The Blueprint QA Multi-Task model scores similarly, achieving nearly identical scores as the baseline.
This finding suggests that the methods generate summaries with similar quality, as measured by ROUGE.

% E2E
The same trend does not hold for E2E methods, which see a steep drop in ROUGE scores.
% In the most extreme case, the Coarse Plans + QA E2E model scores \tocheck{13.2, 4.7 and 4.0 lower} than the Multi-task counterpart across ROUGE-1, -2 and -L on SQuALITY, and 2.75, 1.4 and 1.4 across ROUGE-1, -2 and -L on SummScreen.
This likely occurs due to the difficulty of generating much longer text in a single pass, as observed by \citet{conditional-narayan-etal-2023}.

% Coarse vs Coarse + QA vs. Blueprint QA
Across different plan formulations, we observe very little change in ROUGE performance within each task setting (e.g. across E2E methods or across multi-task methods).
Instead, these methods generally perform similarly to the baseline on both datasets, suggesting that planning does not significantly improve summary quality.
% We note a slight decrease in performance with the Coarse Plans + QA style plans in both E2E and multi-task settings, indicating the complex planning style may have a mildly detrimental effect on summarization.

% Claude baseline
Lastly, the Claude 3.5 Sonnet baseline performs better than E2E methods but worse than other fine-tuned methods on SQuALITY, despite %likely being much larger than Phi-3.5-mini.
undoubtedly being much larger than Phi-3.5-mini.
Claude 3.5 also performs average on SummScreen, achieving the lowest ROUGE-L score overall and lower than the Phi-3 baseline on ROUGE-1 and -2.
Fine-tuning likely confers advantages in terms of either summary quality or in-domain knowledge (for example, the expected summary style).

% The exception to this finding is the Blueprint QA models, which we find generally underperform in their summary quality.
% For example, our \textit{Coarse Plans, Multi-Task} model scores 2.45 ROUGE-1, 0.9 ROUGE-2 and 1.07 ROUGE-L higher than the Blueprint Multi-Task counterpart.
% The result suggests that sub-event modeling offers an advantage in summary quality over fine-grained methods such as QA-only plans in the short story domain.

% We also note a trend that multi-task settings outperform E2E methods in terms of ROUGE.
% With \textit{Coarse Plans}, the multi-task setting outperforms the E2E method by 2.1, 1.21, and 0.31 across ROUGE-1, -2 and -L metrics.
% Similarly, using \textit{Coarse Plans + QA}, the multi-task setting beats E2E by 0.91 and 0.64 on ROUGE-2 and -L metrics, though it trails by 1.24 on ROUGE-1.
% We also see a similar trend in our implementation of the Blueprint QA baselines.
% This improvement validates the hypothesis from \cite{conditional-narayan-etal-2023} about the advantages of multi-task learning: separately learning how to plan and summarize can results in stronger summary quality, as the model can more easily focus on a single generation task.

% Faithfulness
\paragraph{Faithfulness}
We see particularly mixed results across the faithfulness metrics. 
While Claude 3.5 Sonnet achieves the highest AlignScore values on both datasets, it also earns a much lower QAFactEval score, notably lower than the Phi-3.5 baseline.

Likewise, the Coarse Plans + QA, Multi-Task setting achieves the highest QAFactEval scores on SQuALITY, but the Phi-3.5 baseline outperforms other planning methods. 
On SummScreen, all methods either perform comparably or lower than the baseline, and QAFactEval does not suggest that planning improves faithfulness.

Finally, Sonnet 3.5's FineSurE score eclipses other models, achieving 24.4 points over the next highest model on SQuALITY and 40 points on SummScreen.
However, we are wary of this result after previous work has found LLM auto-raters tend to favor their own outputs \cite{kim2024fables}.
Other settings achieve similar scores, except for the Blueprint QA, E2E setting on SummScreen.
As we will note in the following section, Blueprint QA, E2E summaries tend to be drastically short, which we suspect inflates the FineSurE score.

In general, the mixed results lead us to question the reliability of existing faithfulness metrics and whether any method is truly outperforming the baseline. 
This issue motivates our human evaluation in Section \ref{subsec:human-eval}.

\subsection{Human Evaluation}
\label{subsec:human-eval}

Human evaluation results are shown in Table \ref{table:manual-results}.
We note E2E coverage scores tend to be lower than their Multi-Task counterparts. 
We observe that the E2E model occasionally outputs a lengthy plan, leaving little space for the resulting summary before reaching the maximum token count.
This behavior negatively affects coverage as the shortened summary cannot cover all relevant story points.
In the Blueprint QA E2E model, we notice a particular failure case where the model's summary is a nearly identical copy of the QA pairs, resulting in summaries with especially low coverage.

We observe that faithfulness and conciseness scores are roughly equal across all settings. 
Faithfulness scores range from 68 to 77\%, with the highest faithfulness score achieved by the Blueprint QA, Multi-Task setting.
Although we do not report faithfulness scores on the Multi-Task plans, we note these plans also contain hallucinations at roughly the same rate as the E2E plans.
Similarly, conciseness scores range from 72 to 88\%, though we note that the Blueprint QA E2E model scores more than 10 points lower than any other setting.

On grounding, we observe that E2E methods based on Coarse plans are remarkably tightly grounded to their plans, achieving above 90\%.
This effect contrasts with multi-task settings where grounding scores range from 14 to 71\%.
The effect can be readily explained by the different task setups: E2E models directly use their plans via the decoder's attention mechanism, whereas multi-task models only generate plans during training.

\begin{table*}[t]
\small
\centering
\begin{tabular}{l r r r r}
\toprule									
Model	&	Coverage	&	Faithfulness	&	Conciseness	&	Grounding	\\
\midrule									
Phi-3.5-mini &	46.82	&	75.33	&	84.45	&	-	\vspace{0.15cm} \\
Blueprint QA, E2E	&	36.82	&	67.99	&	72.27	&	70.00	\\
Blueprint QA, Multi-Task	&	56.36	&	\textbf{78.41}	&	82.69	&	14.00\vspace{0.15cm} \\
Coarse Plans, E2E	&	43.15	&	74.09	&	84.95	&	91.68	\\
Coarse Plans, Multi-Task	&	57.01	&	74.15	&	\textbf{88.62}	&	71.92\vspace{0.15cm} \\
Coarse Plans + QA, E2E	&	37.71	&	71.26	&	87.05	&	\textbf{97.78}	\\
Coarse Plans + QA, Multi-Task	&	\textbf{64.10}	&	77.39	&	82.81	&	64.79\vspace{0.15cm} \\
Pre-filled Claude Coarse Plans, E2E	&	74.19	&	84.57	&	92.10	&	87.77	\\
\bottomrule
\end{tabular}
\caption{Human evaluation results on SQuALITY. We bold the highest score in each column, excluding the ``Pre-filled Claude Coarse Plans, E2E'' oracle setting.}
\label{table:manual-results}
\end{table*}

\section{Analysis}
% \begin{table}[t]
%     \centering
%     \begin{tabular}{l r r r r r}
% \toprule
% Model	&	R1	&	R2	&	RL	&	AS	&	QF	\\
% \midrule											
% LongT5	&	34.8	&	7.7	&	19.2	&	37.6	&	1.34\vspace{0.15cm}\\
% BQA, E2E	&	35.0	&	6.9	&	18.4	&	35.3	&	0.98	\\
% BQA, MT	&	\textbf{35.6}	&	\textbf{8.6}	&	\textbf{19.5}	&	33.9	&	1.08\vspace{0.15cm}\\
% CP, E2E	&	31.2	&	5.9	&	17.4	&	36.2	&	1.21	\\
% CP, MT	&	35.0	&	7.9	&	19.2	&	36.8	&	\textbf{1.37}\vspace{0.15cm}\\
% CP+QA,E2E	&	32.6	&	5.7	&	17.5	&	36.2	&	1.09	\\
% CP+QA,MT	&	35.2	&	8.0	&	19.4	&	\textbf{38.2}	&	1.22	\\
% \bottomrule	
%     \end{tabular}
%     \caption{Results on a subset of SummScreen where summaries with less than three sentences are dropped. BQA=Blueprint QA, CP=Coarse Plans, MT=Multi-Task, AS=AlignScore, QF=QAFactEval.}
%     \label{tab:summscreen-drop}
% \end{table}

\subsection{Claude Synthetic Plans}
\begin{table}[t]
    \small
    \centering
    \begin{tabular}{l c c c c}
    \toprule
    Dataset & Cov. & Faith. & Red. & Cit. Acc. \\
    \midrule
    SQuALITY & 84.44 & 98.26 & 0.87 & 91.26 \\
    SummScreen & 98.67 & 98.63 & 0.0 & 100.0 \\
    \bottomrule
    \end{tabular}
    \caption{Analysis of coarse plans generated by Claude Sonnet 3.5 v1. Cov.=Coverage, Faith.=Faithfulness, Red.=Redundancy, and Cit. Acc.=Citation Accuracy.}
    \label{tab:claude-plan-analysis}
\end{table}
We would like to know if plans generated by Claude 3.5 (plans we train on) correctly correspond to relevant sub-events in text. %reasonably correspond to sub-events in the text as expected.
To this end, we manually inspect 20 coarse plans generated by Claude 3.5 on SummScreen and SQuALITY, covering 115 generated plan points on SQuALITY and 73 on SummScreen.
Similar to our human evaluation, we analyze if each generated plan point is factual to the source summary (faithfulness), and whether each key fact in the reference summary is contained in the generated plan (coverage). 
% Since summaries contain many facts which are not essential to the story and not expected to be presented in the coarse plan, we only analyze key facts from the summary and ask if they are present in the generated plan.
We inspect each plan for redundant plan points, and report the percentage of redundant points among all generated points.
Finally, we verify the accuracy of citations by checking if each citation provides a suitable source sentence for the associated plan point.

The results are shown in Table \ref{tab:claude-plan-analysis}.
In general, we find that Claude's plans are very high quality.
Across both datasets, faithfulness scores are above 98\%, and we find that Claude rarely includes hallucinated details in the coarse plan.

Coverage performance is similarly high on SummScreen but lower on SQuALITY.
We find that the summary length has a significant effect on Claude's ability to extract key events, and on longer summaries, Claude's plans tend to omit key details more often.
On SummScreen, where summaries are generally shorter, this effect is less noticeable.

On both datasets, Claude's plans are highly non-redundant -- we find only one case of redundant plan points, in SQuALITY. Likewise, the generated citations are highly accurate, scoring above 90\% on SQuALITY and flawlessly on SummScreen.
% Overall, we see that Claude's generated coarse plans are high quality, with some difficulties extracting key facts on longer summaries in SQuALITY.

\subsection{Pre-filled Claude Plans in E2E Setting}
Our results have shown that that E2E settings tend to be very grounded in their plans, but factual errors in planning ultimately lead to unfaithful summaries.
On the other hand, Claude 3.5 plans are much higher in quality than the predicted plans across all of our tested settings. 
Putting the two together, we are interested to know whether substituting Claude Sonnet 3.5 plans for the predicted ones could lead to better summaries overall.
Although such an implementation would be impractical in a real world setting, it can give insight into whether high-quality plans can lead to better summaries.

% method
We run the Coarse Plans E2E settings as before (without QA) on SQuALITY, but replace their predicted plans with synthetic plans Claude 3.5 plans.
We then score the generated summaries as before.

% automatic results
% large increase in rouge scores
% this is most likely because the plans are provided and do not encroach on the summaries
The automatic evaluation results are shown in Table \ref{table:results} (bottom row of SQuALITY section).
Using the pre-filled plans greatly improves ROUGE scores compared to the normal E2E settings.
Although it is tempting to draw the conclusion that the summaries are therefore higher quality, we suspect this effect is mainly because E2E settings tend to overgenerate plans, crowding out their summaries.
Since the plans are now pre-filled, E2E models are no longer at risk of generating excessively lengthy plans.
Faithfulness scores on AlignScore and QAFactEval decrease slightly, though we again question the reliability of these metrics.

% manual results
Our manual evaluation results are shown in Table \ref{table:manual-results}. 
E2E with Claude plans maintains good grounding to plans as before, although with a slight decrease to other E2E methods.
We observe a notable increase in coverage (+10 over the next highest), as the Claude plans' coverage is much higher than compared to other planning models.
For the same reason, conciseness scores also increase, as Claude plans are typically concise.

Faithfulness also slightly increases by 6.16\% over the next highest score.
We notice that although planning content is highly factual, the E2E model tends to follow the plan but then veer off into hallucinated generation.
Overall, pre-filling plans increases scores across coverage, faithfulness and conciseness scores.
Of course, the main challenge is generating very high quality plans without resorting to an LLM.%. such as Claude Sonnet 3.5 (v1).

% Our proposed methods generally perform worse on SummScreen compared to SQuALITY, in particular on  AlignScore where the LongT5 baseline scores greater than any method with planning. 
% Many examples in SummScreen contain summaries with less than three sentences, and appear to be more similar to synopses rather than full summaries.
% These short summaries occur in 28.3\%, 22.4\% and 22.8\% of training, validation and test examples respectively.

% Along with the issues with AlignScore discussed previously, we hypothesize that in cases with very short summaries, planning serves little use.
% We re-train and re-evaluate our models after dropping examples with summaries of less than three sentences in SummScreen.

% Table \ref{tab:summscreen-drop} shows the results for this experiment. %The results show 
% Dropping examples with short summaries benefits our proposed planning methods, particularly the \textit{Coarse Plans+QA, Multi-Task} setting. 
% This model maintains similar ROUGE scores compared to the baseline, and achieves higher faithfulness scores: +0.6 AlignScore and +0.17 QAFactEval.
% We note that higher-level planning is more effective in tasks with longer summaries.
% These examples are more likely to include narrative structure compared to synopsis-like summaries.

% In contrast, Blueprint QA methods generally see increases in ROUGE scores but decreases in faithfulness metrics.
% These methods may benefit from comparison to shorter summaries as they are less likely to detail the narrative structure in the source text that our methods are designed to handle.

\section{Conclusion}
In this work, we investigate several plan formulations for plan-guided summarization in narrative texts.
We explore lower-level plans with targeted QA pairs, higher-level plans that reflect the narrative structure of the text, and a combination of both.
Our high-level planning method involves prompting an LLM to produce training plans, followed by fine-tuning a smaller decoder model.
Our results show that, contrary to prior works, planning methods do not offer a considerable improvement in summary quality or faithfulness.
In our manual evaluation, we find that while E2E methods are fairly grounded to summaries, all settings hallucinate at similar rates, including in the generated plans.
Replacing E2E plans with high-quality Claude plans improves summary quality, but only partially mitigates faithfulness issues. %, but issues with faithfulness persist.
Our work illustrates the difficulty of plan-guided summarization in narrative text with small LMs.

\section{Limitations}
% As noted throughout this work, models with or without planning generate factual errors in their summaries. 
Although we use a variety of faithfulness metrics throughout this work, we note that they often do not correlate and may even contradict each other. 
These scores should be interpreted with caution, and we include manual evaluation to provide additional perspectives on models' faithfulness.
Future work could consider more scalable yet reliable evaluation methods, in particular using different LLMs as auto-raters in the FineSurE metric.
% In this work, we explore high-level planning as a method to reduce factual errors in summarization, but our final models may still make serious errors in generation. 
% As mentioned previously, the AlignScore may not be reliable in every domain, and even high-scoring models may still make factual mistakes. 
% Users interested in plan-guided summarization methods such as this one should be careful to analyze for potentially false or even harmful expressions in the model outputs.

In prior work, \citet{conditional-narayan-etal-2023} also propose an \textit{iterative} method, where question-answer pairs and summary sentences are interwoven; at each iteration, the model first predicts a question-answer pair and then conditions of this plan to generate a sentence summary.
This method is attractive because it allows for controllability: question-answer pairs can be edited, resulting in a new summary more suited to the user's needs.
They show the iterative mechanism also promotes higher faithfulness scores compared to other settings.
We looked at implementing such an iterative setting using coarse plans and summary sentences; however, the method scored poorly and we subsequently focused E2E and multi-task settings instead.
The methods we propose are therefore not as flexible as other works that support iterative methods.
In the future, such a method would allow a more controllable model, as sub-events in predicted plans could be inspected and modified, resulting in a summary that reflects an alternative narrative structure.

\section*{Acknowledgements}

We would like to thank Faisal Ladhak and Kathleen McKeown for their helpful discussion early in the project.

\bibliography{anthology_external,custom}

\begin{thebibliography}{38}
\providecommand{\natexlab}[1]{#1}

\bibitem[{Abdin et~al.(2024)Abdin, Aneja, Awadalla, Awadallah, Awan, Bach, Bahree, Bakhtiari, Bao, Behl, Benhaim, Bilenko, Bjorck, Bubeck, Cai, Cai, Chaudhary, Chen, Chen, Chen, Chen, Chen, Cheng, Chopra, Dai, Dixon, Eldan, Fragoso, Gao, Gao, Gao, Garg, Giorno, Goswami, Gunasekar, Haider, Hao, Hewett, Hu, Huynh, Iter, Jacobs, Javaheripi, Jin, Karampatziakis, Kauffmann, Khademi, Kim, Kim, Kurilenko, Lee, Lee, Li, Li, Liang, Liden, Lin, Lin, Liu, Liu, Liu, Liu, Liu, Luo, Madan, Mahmoudzadeh, Majercak, Mazzola, Mendes, Mitra, Modi, Nguyen, Norick, Patra, Perez-Becker, Portet, Pryzant, Qin, Radmilac, Ren, de~Rosa, Rosset, Roy, Ruwase, Saarikivi, Saied, Salim, Santacroce, Shah, Shang, Sharma, Shen, Shukla, Song, Tanaka, Tupini, Vaddamanu, Wang, Wang, Wang, Wang, Wang, Wang, Ward, Wen, Witte, Wu, Wu, Wyatt, Xiao, Xu, Xu, Xu, Xue, Yadav, Yang, Yang, Yang, Yang, Yu, Yuan, Zhang, Zhang, Zhang, Zhang, Zhang, Zhang, Zhang, and Zhou}]{phi3}
Marah Abdin, Jyoti Aneja, Hany Awadalla, Ahmed Awadallah, Ammar~Ahmad Awan, Nguyen Bach, Amit Bahree, Arash Bakhtiari, Jianmin Bao, Harkirat Behl, Alon Benhaim, Misha Bilenko, Johan Bjorck, Sébastien Bubeck, Martin Cai, Qin Cai, Vishrav Chaudhary, Dong Chen, Dongdong Chen, Weizhu Chen, Yen-Chun Chen, Yi-Ling Chen, Hao Cheng, Parul Chopra, Xiyang Dai, Matthew Dixon, Ronen Eldan, Victor Fragoso, Jianfeng Gao, Mei Gao, Min Gao, Amit Garg, Allie~Del Giorno, Abhishek Goswami, Suriya Gunasekar, Emman Haider, Junheng Hao, Russell~J. Hewett, Wenxiang Hu, Jamie Huynh, Dan Iter, Sam~Ade Jacobs, Mojan Javaheripi, Xin Jin, Nikos Karampatziakis, Piero Kauffmann, Mahoud Khademi, Dongwoo Kim, Young~Jin Kim, Lev Kurilenko, James~R. Lee, Yin~Tat Lee, Yuanzhi Li, Yunsheng Li, Chen Liang, Lars Liden, Xihui Lin, Zeqi Lin, Ce~Liu, Liyuan Liu, Mengchen Liu, Weishung Liu, Xiaodong Liu, Chong Luo, Piyush Madan, Ali Mahmoudzadeh, David Majercak, Matt Mazzola, Caio César~Teodoro Mendes, Arindam Mitra, Hardik Modi, Anh Nguyen,
  Brandon Norick, Barun Patra, Daniel Perez-Becker, Thomas Portet, Reid Pryzant, Heyang Qin, Marko Radmilac, Liliang Ren, Gustavo de~Rosa, Corby Rosset, Sambudha Roy, Olatunji Ruwase, Olli Saarikivi, Amin Saied, Adil Salim, Michael Santacroce, Shital Shah, Ning Shang, Hiteshi Sharma, Yelong Shen, Swadheen Shukla, Xia Song, Masahiro Tanaka, Andrea Tupini, Praneetha Vaddamanu, Chunyu Wang, Guanhua Wang, Lijuan Wang, Shuohang Wang, Xin Wang, Yu~Wang, Rachel Ward, Wen Wen, Philipp Witte, Haiping Wu, Xiaoxia Wu, Michael Wyatt, Bin Xiao, Can Xu, Jiahang Xu, Weijian Xu, Jilong Xue, Sonali Yadav, Fan Yang, Jianwei Yang, Yifan Yang, Ziyi Yang, Donghan Yu, Lu~Yuan, Chenruidong Zhang, Cyril Zhang, Jianwen Zhang, Li~Lyna Zhang, Yi~Zhang, Yue Zhang, Yunan Zhang, and Xiren Zhou. 2024.
\newblock \href {https://arxiv.org/abs/2404.14219} {Phi-3 technical report: A highly capable language model locally on your phone}.
\newblock \emph{Preprint}, arXiv:2404.14219.

\bibitem[{Adams et~al.(2023)Adams, Fabbri, Ladhak, Elhadad, and McKeown}]{adams-etal-2023-generating}
Griffin Adams, Alex Fabbri, Faisal Ladhak, No{\'e}mie Elhadad, and Kathleen McKeown. 2023.
\newblock \href {https://doi.org/10.18653/v1/2023.acl-long.151} {Generating {EDU} extracts for plan-guided summary re-ranking}.
\newblock In \emph{Proceedings of the 61st Annual Meeting of the Association for Computational Linguistics (Volume 1: Long Papers)}, pages 2680--2697, Toronto, Canada. Association for Computational Linguistics.

\bibitem[{Anthropic(2024)}]{claude3}
Anthropic. 2024.
\newblock \href {https://www-cdn.anthropic.com/fed9cc193a14b84131812372d8d5857f8f304c52/Model_Card_Claude_3_Addendum.pdf} {Claude 3.5 sonnet model card addendum}.
\newblock Accessed on October 3, 2024.

\bibitem[{Brown et~al.(2020)Brown, Mann, Ryder, Subbiah, Kaplan, Dhariwal, Neelakantan, Shyam, Sastry, Askell, Agarwal, Herbert-Voss, Krueger, Henighan, Child, Ramesh, Ziegler, Wu, Winter, Hesse, Chen, Sigler, Litwin, Gray, Chess, Clark, Berner, McCandlish, Radford, Sutskever, and Amodei}]{gpt3}
Tom Brown, Benjamin Mann, Nick Ryder, Melanie Subbiah, Jared~D Kaplan, Prafulla Dhariwal, Arvind Neelakantan, Pranav Shyam, Girish Sastry, Amanda Askell, Sandhini Agarwal, Ariel Herbert-Voss, Gretchen Krueger, Tom Henighan, Rewon Child, Aditya Ramesh, Daniel Ziegler, Jeffrey Wu, Clemens Winter, Chris Hesse, Mark Chen, Eric Sigler, Mateusz Litwin, Scott Gray, Benjamin Chess, Jack Clark, Christopher Berner, Sam McCandlish, Alec Radford, Ilya Sutskever, and Dario Amodei. 2020.
\newblock \href {https://proceedings.neurips.cc/paper_files/paper/2020/file/1457c0d6bfcb4967418bfb8ac142f64a-Paper.pdf} {Language models are few-shot learners}.
\newblock In \emph{Advances in Neural Information Processing Systems}, volume~33, pages 1877--1901. Curran Associates, Inc.

\bibitem[{Chawla et~al.(2024)Chawla, Rashkin, Tomar, and Reitter}]{chawla-etal-2024-investigating}
Kushal Chawla, Hannah Rashkin, Gaurav~Singh Tomar, and David Reitter. 2024.
\newblock \href {https://aclanthology.org/2024.eacl-long.142/} {Investigating content planning for navigating trade-offs in knowledge-grounded dialogue}.
\newblock In \emph{Proceedings of the 18th Conference of the European Chapter of the Association for Computational Linguistics (Volume 1: Long Papers)}, pages 2316--2335, St. Julian{'}s, Malta. Association for Computational Linguistics.

\bibitem[{Chen et~al.(2022)Chen, Chu, Wiseman, and Gimpel}]{chen-etal-2022-summscreen}
Mingda Chen, Zewei Chu, Sam Wiseman, and Kevin Gimpel. 2022.
\newblock \href {https://doi.org/10.18653/v1/2022.acl-long.589} {{S}umm{S}creen: A dataset for abstractive screenplay summarization}.
\newblock In \emph{Proceedings of the 60th Annual Meeting of the Association for Computational Linguistics (Volume 1: Long Papers)}, pages 8602--8615, Dublin, Ireland. Association for Computational Linguistics.

\bibitem[{Daniel et~al.(2003)Daniel, Radev, and Allison}]{daniel-etal-2003-sub}
Naomi Daniel, Dragomir Radev, and Timothy Allison. 2003.
\newblock \href {https://aclanthology.org/W03-0502/} {Sub-event based multi-document summarization}.
\newblock In \emph{Proceedings of the {HLT}-{NAACL} 03 Text Summarization Workshop}, pages 9--16.

\bibitem[{Deutsch and Roth(2023)}]{deutsch-roth-2023-incorporating}
Daniel Deutsch and Dan Roth. 2023.
\newblock \href {https://doi.org/10.18653/v1/2023.eacl-main.42} {Incorporating question answering-based signals into abstractive summarization via salient span selection}.
\newblock In \emph{Proceedings of the 17th Conference of the European Chapter of the Association for Computational Linguistics}, pages 575--588, Dubrovnik, Croatia. Association for Computational Linguistics.

\bibitem[{Fabbri et~al.(2022)Fabbri, Wu, Liu, and Xiong}]{fabbri-etal-2022-qafacteval}
Alexander Fabbri, Chien-Sheng Wu, Wenhao Liu, and Caiming Xiong. 2022.
\newblock \href {https://doi.org/10.18653/v1/2022.naacl-main.187} {{QAF}act{E}val: Improved {QA}-based factual consistency evaluation for summarization}.
\newblock In \emph{Proceedings of the 2022 Conference of the North American Chapter of the Association for Computational Linguistics: Human Language Technologies}, pages 2587--2601, Seattle, United States. Association for Computational Linguistics.

\bibitem[{Fierro et~al.(2024)Fierro, Amplayo, Huot, De~Cao, Maynez, Narayan, and Lapata}]{fierro-etal-2024-learning}
Constanza Fierro, Reinald~Kim Amplayo, Fantine Huot, Nicola De~Cao, Joshua Maynez, Shashi Narayan, and Mirella Lapata. 2024.
\newblock \href {https://doi.org/10.18653/v1/2024.acl-long.615} {Learning to plan and generate text with citations}.
\newblock In \emph{Proceedings of the 62nd Annual Meeting of the Association for Computational Linguistics (Volume 1: Long Papers)}, pages 11397--11417, Bangkok, Thailand. Association for Computational Linguistics.

\bibitem[{Gao et~al.(2020)Gao, Biderman, Black, Golding, Hoppe, Foster, Phang, He, Thite, Nabeshima, Presser, and Leahy}]{gao2020pile}
Leo Gao, Stella Biderman, Sid Black, Laurence Golding, Travis Hoppe, Charles Foster, Jason Phang, Horace He, Anish Thite, Noa Nabeshima, Shawn Presser, and Connor Leahy. 2020.
\newblock \href {https://arxiv.org/abs/2101.00027} {The pile: An 800gb dataset of diverse text for language modeling}.
\newblock \emph{Preprint}, arXiv:2101.00027.

\bibitem[{Glikson and Woolley(2020)}]{hci1}
Ella Glikson and Anita~Williams Woolley. 2020.
\newblock \href {https://doi.org/10.5465/annals.2018.0057} {Human trust in artificial intelligence: Review of empirical research}.
\newblock \emph{Academy of Management Annals}, 14(2):627--660.

\bibitem[{Godbole et~al.(2024)Godbole, Monath, Kim, Rawat, McCallum, and Zaheer}]{godbole-etal-2024-analysis}
Ameya Godbole, Nicholas Monath, Seungyeon Kim, Ankit~Singh Rawat, Andrew McCallum, and Manzil Zaheer. 2024.
\newblock \href {https://doi.org/10.18653/v1/2024.emnlp-main.727} {Analysis of plan-based retrieval for grounded text generation}.
\newblock In \emph{Proceedings of the 2024 Conference on Empirical Methods in Natural Language Processing}, pages 13101--13119, Miami, Florida, USA. Association for Computational Linguistics.

\bibitem[{Gunjal and Durrett(2024)}]{gunjal-durrett-2024-molecular}
Anisha Gunjal and Greg Durrett. 2024.
\newblock \href {https://doi.org/10.18653/v1/2024.findings-emnlp.215} {Molecular facts: Desiderata for decontextualization in {LLM} fact verification}.
\newblock In \emph{Findings of the Association for Computational Linguistics: EMNLP 2024}, pages 3751--3768, Miami, Florida, USA. Association for Computational Linguistics.

\bibitem[{Guo et~al.(2022)Guo, Ainslie, Uthus, Ontanon, Ni, Sung, and Yang}]{guo-etal-2022-longt5}
Mandy Guo, Joshua Ainslie, David Uthus, Santiago Ontanon, Jianmo Ni, Yun-Hsuan Sung, and Yinfei Yang. 2022.
\newblock \href {https://doi.org/10.18653/v1/2022.findings-naacl.55} {{L}ong{T}5: {E}fficient text-to-text transformer for long sequences}.
\newblock In \emph{Findings of the Association for Computational Linguistics: NAACL 2022}, pages 724--736, Seattle, United States. Association for Computational Linguistics.

\bibitem[{Honnibal et~al.(2020)Honnibal, Montani, Van~Landeghem, and Boyd}]{spacy}
Matthew Honnibal, Ines Montani, Sofie Van~Landeghem, and Adriane Boyd. 2020.
\newblock \href {https://doi.org/10.5281/zenodo.1212303} {{spaCy: Industrial-strength Natural Language Processing in Python}}.

\bibitem[{Hu et~al.(2022)Hu, Shen, Wallis, Allen-Zhu, Li, Wang, Wang, and Chen}]{hu2022lora}
Edward~J Hu, Yelong Shen, Phillip Wallis, Zeyuan Allen-Zhu, Yuanzhi Li, Shean Wang, Lu~Wang, and Weizhu Chen. 2022.
\newblock \href {https://openreview.net/forum?id=nZeVKeeFYf9} {Lo{RA}: Low-rank adaptation of large language models}.
\newblock In \emph{International Conference on Learning Representations}.

\bibitem[{Hua et~al.(2023)Hua, Deng, and McKeown}]{hua-etal-2023-improving}
Yilun Hua, Zhaoyuan Deng, and Kathleen McKeown. 2023.
\newblock \href {https://doi.org/10.18653/v1/2023.findings-acl.871} {Improving long dialogue summarization with semantic graph representation}.
\newblock In \emph{Findings of the Association for Computational Linguistics: ACL 2023}, pages 13851--13883, Toronto, Canada. Association for Computational Linguistics.

\bibitem[{Ji et~al.(2023)Ji, Lee, Frieske, Yu, Su, Xu, Ishii, Bang, Madotto, and Fung}]{ziwei_etal_2023}
Ziwei Ji, Nayeon Lee, Rita Frieske, Tiezheng Yu, Dan Su, Yan Xu, Etsuko Ishii, Ye~Jin Bang, Andrea Madotto, and Pascale Fung. 2023.
\newblock \href {https://doi.org/10.1145/3571730} {Survey of hallucination in natural language generation}.
\newblock \emph{ACM Comput. Surv.}, 55(12).

\bibitem[{Kim et~al.(2024)Kim, Chang, Karpinska, Garimella, Manjunatha, Lo, Goyal, and Iyyer}]{kim2024fables}
Yekyung Kim, Yapei Chang, Marzena Karpinska, Aparna Garimella, Varun Manjunatha, Kyle Lo, Tanya Goyal, and Mohit Iyyer. 2024.
\newblock \href {https://openreview.net/forum?id=YfHxQSoaWU} {{FABLES}: Evaluating faithfulness and content selection in book-length summarization}.
\newblock In \emph{First Conference on Language Modeling}.

\bibitem[{Kryscinski et~al.(2019)Kryscinski, Keskar, McCann, Xiong, and Socher}]{kryscinski-etal-2019-neural}
Wojciech Kryscinski, Nitish~Shirish Keskar, Bryan McCann, Caiming Xiong, and Richard Socher. 2019.
\newblock \href {https://doi.org/10.18653/v1/D19-1051} {Neural text summarization: A critical evaluation}.
\newblock In \emph{Proceedings of the 2019 Conference on Empirical Methods in Natural Language Processing and the 9th International Joint Conference on Natural Language Processing (EMNLP-IJCNLP)}, pages 540--551, Hong Kong, China. Association for Computational Linguistics.

\bibitem[{Kryscinski et~al.(2022)Kryscinski, Rajani, Agarwal, Xiong, and Radev}]{kryscinski-etal-2022-booksum}
Wojciech Kryscinski, Nazneen Rajani, Divyansh Agarwal, Caiming Xiong, and Dragomir Radev. 2022.
\newblock \href {https://doi.org/10.18653/v1/2022.findings-emnlp.488} {{BOOKSUM}: A collection of datasets for long-form narrative summarization}.
\newblock In \emph{Findings of the Association for Computational Linguistics: EMNLP 2022}, pages 6536--6558, Abu Dhabi, United Arab Emirates. Association for Computational Linguistics.

\bibitem[{Levy et~al.(2024)Levy, Jacoby, and Goldberg}]{levy-etal-2024-task}
Mosh Levy, Alon Jacoby, and Yoav Goldberg. 2024.
\newblock \href {https://doi.org/10.18653/v1/2024.acl-long.818} {Same task, more tokens: the impact of input length on the reasoning performance of large language models}.
\newblock In \emph{Proceedings of the 62nd Annual Meeting of the Association for Computational Linguistics (Volume 1: Long Papers)}, pages 15339--15353, Bangkok, Thailand. Association for Computational Linguistics.

\bibitem[{Lewis et~al.(2020)Lewis, Liu, Goyal, Ghazvininejad, Mohamed, Levy, Stoyanov, and Zettlemoyer}]{lewis-etal-2020-bart}
Mike Lewis, Yinhan Liu, Naman Goyal, Marjan Ghazvininejad, Abdelrahman Mohamed, Omer Levy, Veselin Stoyanov, and Luke Zettlemoyer. 2020.
\newblock \href {https://doi.org/10.18653/v1/2020.acl-main.703} {{BART}: Denoising sequence-to-sequence pre-training for natural language generation, translation, and comprehension}.
\newblock In \emph{Proceedings of the 58th Annual Meeting of the Association for Computational Linguistics}, pages 7871--7880, Online. Association for Computational Linguistics.

\bibitem[{Lin(2004)}]{lin-2004-rouge}
Chin-Yew Lin. 2004.
\newblock \href {https://aclanthology.org/W04-1013/} {{ROUGE}: A package for automatic evaluation of summaries}.
\newblock In \emph{Text Summarization Branches Out}, pages 74--81, Barcelona, Spain. Association for Computational Linguistics.

\bibitem[{Lin et~al.(2022)Lin, Hilton, and Evans}]{lin-etal-2022-truthfulqa}
Stephanie Lin, Jacob Hilton, and Owain Evans. 2022.
\newblock \href {https://doi.org/10.18653/v1/2022.acl-long.229} {{T}ruthful{QA}: Measuring how models mimic human falsehoods}.
\newblock In \emph{Proceedings of the 60th Annual Meeting of the Association for Computational Linguistics (Volume 1: Long Papers)}, pages 3214--3252, Dublin, Ireland. Association for Computational Linguistics.

\bibitem[{Liu et~al.(2024)Liu, Lin, Hewitt, Paranjape, Bevilacqua, Petroni, and Liang}]{liu-etal-2024-lost}
Nelson~F. Liu, Kevin Lin, John Hewitt, Ashwin Paranjape, Michele Bevilacqua, Fabio Petroni, and Percy Liang. 2024.
\newblock \href {https://doi.org/10.1162/tacl_a_00638} {Lost in the middle: How language models use long contexts}.
\newblock \emph{Transactions of the Association for Computational Linguistics}, 12:157--173.

\bibitem[{Maynez et~al.(2020)Maynez, Narayan, Bohnet, and McDonald}]{maynez-etal-2020-faithfulness}
Joshua Maynez, Shashi Narayan, Bernd Bohnet, and Ryan McDonald. 2020.
\newblock \href {https://doi.org/10.18653/v1/2020.acl-main.173} {On faithfulness and factuality in abstractive summarization}.
\newblock In \emph{Proceedings of the 58th Annual Meeting of the Association for Computational Linguistics}, pages 1906--1919, Online. Association for Computational Linguistics.

\bibitem[{Murakhovs{'}ka et~al.(2022)Murakhovs{'}ka, Wu, Laban, Niu, Liu, and Xiong}]{murakhovska-etal-2022-mixqg}
Lidiya Murakhovs{'}ka, Chien-Sheng Wu, Philippe Laban, Tong Niu, Wenhao Liu, and Caiming Xiong. 2022.
\newblock \href {https://doi.org/10.18653/v1/2022.findings-naacl.111} {{M}ix{QG}: Neural question generation with mixed answer types}.
\newblock In \emph{Findings of the Association for Computational Linguistics: NAACL 2022}, pages 1486--1497, Seattle, United States. Association for Computational Linguistics.

\bibitem[{Narayan et~al.(2023{\natexlab{a}})Narayan, Maynez, Amplayo, Ganchev, Louis, Huot, Sandholm, Das, and Lapata}]{conditional-narayan-etal-2023}
Shashi Narayan, Joshua Maynez, Reinald~Kim Amplayo, Kuzman Ganchev, Annie Louis, Fantine Huot, Anders Sandholm, Dipanjan Das, and Mirella Lapata. 2023{\natexlab{a}}.
\newblock \href {https://doi.org/10.1162/tacl_a_00583} {{Conditional Generation with a Question-Answering Blueprint}}.
\newblock \emph{Transactions of the Association for Computational Linguistics}, 11:974--996.

\bibitem[{Narayan et~al.(2023{\natexlab{b}})Narayan, Maynez, Amplayo, Ganchev, Louis, Huot, Sandholm, Das, and Lapata}]{narayan-etal-2023-conditional}
Shashi Narayan, Joshua Maynez, Reinald~Kim Amplayo, Kuzman Ganchev, Annie Louis, Fantine Huot, Anders Sandholm, Dipanjan Das, and Mirella Lapata. 2023{\natexlab{b}}.
\newblock \href {https://doi.org/10.1162/tacl_a_00583} {Conditional generation with a question-answering blueprint}.
\newblock \emph{Transactions of the Association for Computational Linguistics}, 11:974--996.

\bibitem[{Narayan et~al.(2021)Narayan, Zhao, Maynez, Sim{\~o}es, Nikolaev, and McDonald}]{narayan-etal-2021-planning}
Shashi Narayan, Yao Zhao, Joshua Maynez, Gon{\c{c}}alo Sim{\~o}es, Vitaly Nikolaev, and Ryan McDonald. 2021.
\newblock \href {https://doi.org/10.1162/tacl_a_00438} {Planning with learned entity prompts for abstractive summarization}.
\newblock \emph{Transactions of the Association for Computational Linguistics}, 9:1475--1492.

\bibitem[{Pagnoni et~al.(2023)Pagnoni, Fabbri, Kryscinski, and Wu}]{pagnoni-etal-2023-socratic}
Artidoro Pagnoni, Alex Fabbri, Wojciech Kryscinski, and Chien-Sheng Wu. 2023.
\newblock \href {https://doi.org/10.18653/v1/2023.acl-long.713} {Socratic pretraining: Question-driven pretraining for controllable summarization}.
\newblock In \emph{Proceedings of the 61st Annual Meeting of the Association for Computational Linguistics (Volume 1: Long Papers)}, pages 12737--12755, Toronto, Canada. Association for Computational Linguistics.

\bibitem[{Shao et~al.(2024)Shao, Jiang, Kanell, Xu, Khattab, and Lam}]{shao-etal-2024-assisting}
Yijia Shao, Yucheng Jiang, Theodore Kanell, Peter Xu, Omar Khattab, and Monica Lam. 2024.
\newblock \href {https://doi.org/10.18653/v1/2024.naacl-long.347} {Assisting in writing {W}ikipedia-like articles from scratch with large language models}.
\newblock In \emph{Proceedings of the 2024 Conference of the North American Chapter of the Association for Computational Linguistics: Human Language Technologies (Volume 1: Long Papers)}, pages 6252--6278, Mexico City, Mexico. Association for Computational Linguistics.

\bibitem[{Song et~al.(2024)Song, Su, Shalyminov, Cai, and Mansour}]{song-etal-2024-finesure}
Hwanjun Song, Hang Su, Igor Shalyminov, Jason Cai, and Saab Mansour. 2024.
\newblock \href {https://doi.org/10.18653/v1/2024.acl-long.51} {{F}ine{S}ur{E}: Fine-grained summarization evaluation using {LLM}s}.
\newblock In \emph{Proceedings of the 62nd Annual Meeting of the Association for Computational Linguistics (Volume 1: Long Papers)}, pages 906--922, Bangkok, Thailand. Association for Computational Linguistics.

\bibitem[{Wang et~al.(2022)Wang, Pang, Chen, Phang, and Bowman}]{wang-etal-2022-squality}
Alex Wang, Richard~Yuanzhe Pang, Angelica Chen, Jason Phang, and Samuel~R. Bowman. 2022.
\newblock \href {https://doi.org/10.18653/v1/2022.emnlp-main.75} {{SQ}u{ALITY}: Building a long-document summarization dataset the hard way}.
\newblock In \emph{Proceedings of the 2022 Conference on Empirical Methods in Natural Language Processing}, pages 1139--1156, Abu Dhabi, United Arab Emirates. Association for Computational Linguistics.

\bibitem[{Wang et~al.(2023)Wang, Liu, Yue, Tang, Zhang, Jiayang, Yao, Gao, Hu, Qi, Wang, Yang, Wang, Xie, Zhang, and Zhang}]{wang2023survey}
Cunxiang Wang, Xiaoze Liu, Yuanhao Yue, Xiangru Tang, Tianhang Zhang, Cheng Jiayang, Yunzhi Yao, Wenyang Gao, Xuming Hu, Zehan Qi, Yidong Wang, Linyi Yang, Jindong Wang, Xing Xie, Zheng Zhang, and Yue Zhang. 2023.
\newblock \href {https://arxiv.org/abs/2310.07521} {Survey on factuality in large language models: Knowledge, retrieval and domain-specificity}.
\newblock \emph{Preprint}, arXiv:2310.07521.

\bibitem[{Zha et~al.(2023)Zha, Yang, Li, and Hu}]{zha-etal-2023-alignscore}
Yuheng Zha, Yichi Yang, Ruichen Li, and Zhiting Hu. 2023.
\newblock \href {https://doi.org/10.18653/v1/2023.acl-long.634} {{A}lign{S}core: Evaluating factual consistency with a unified alignment function}.
\newblock In \emph{Proceedings of the 61st Annual Meeting of the Association for Computational Linguistics (Volume 1: Long Papers)}, pages 11328--11348, Toronto, Canada. Association for Computational Linguistics.

\end{thebibliography}

\appendix
%%%%%%%%%%%%%% SECTIONS %%%%%%%%%%%%%%%
\section{Examples}\label{sec:examples}
An example of a synthetic plan from Sonnet 3.5, alongside the associated gold summary is shown in Figure \ref{fig:full-plan-example}.
An example of a coarse plan with citations, used for generating QA pairs, is shown in Figure \ref{fig:full-plan-example-w-citations}.

\section{Coarse Planning Prompt}
\label{sec:coarse-prompt}
The prompt used by Sonnet 3.5 to generate plans is shown in Figure \ref{fig:plan-prompt}.

\section{Phi-3.5-mini Baseline Prompt}
\label{sec:phi-baseline-prompt}
The Phi-3.5-mini baseline summarization prompt is shown in Figure \ref{fig:phi3-prompts}.

\section{Claude Baseline Prompt}
\label{sec:claude-baseline-prompt}
The best performing summarization prompt of the 16 summarization prompts we tried is listed in Figure \ref{fig:summ-prompt}.

\section{Compute Budget}
Given that we used a max source text length of 8K, we used a p4d.24xlarge instance available on Amazon Sagemaker which includes 8 NVIDIA A100 GPUs. It costs $\$32.77$ per hour. The training time varied from 4 -- 24 hours for different settings. 

\section{Human Evaluation Details}\label{sec:human-eval-app}
Human evaluation is carried out by two authors of this work.
Periodic discussions were carried out to calibrate and align on edge cases.
Our evaluation rubric can be found in Figure \ref{fig:rubric}, and we provide an example of the type of claims extracted in Figure \ref{fig:claim-example}.

\section{Additional Training Details}\label{sec:training-details}
All models are trained with a learning rate of $0.001$ and a batch size of 32 (per device batch size of 4).
We fine-tune the model using LoRa\footnote{We use $r=24, \alpha=32$ and lora dropout of 0.05} \cite{hu2022lora} for 100 epochs and checkpoint the model using the validation loss.
We truncate input documents to 8K tokens.
Since plans may be long, output length is a key consideration and we perform hyperparameter search over the output generation length with 512, 768 and 1024 tokens.
Accordingly, we set the output length in all E2E settings to 768 and otherwise use 512 tokens.
The E2E setting requires a longer generation size since it generates both plan and summary in a single decoder pass.
During inference, we do not sample, use beam search with 5 beams and a length penalty of $0.8$.

Additionally, for both settings, we forbid repeated ngrams in the generated output for any ngram of length 5, where ngram size is counted as the number of subtoken units. We use this value after running a hyperparameter search over $\{4,..,16\}$ during early experiments.

%%%%%%%%%%%%%% FIGURES %%%%%%%%%%%%%%%
\begin{figure*}[t]
    \centering
    \small
    \begin{tabular}{|p{0.95\linewidth}|}
         \hline
         \vspace{0.1cm}
Captain Linden and his lieutenant "Split" Campbell make up the first manned expedition from Earth to this particular planet, aiming to investigate a large silver river on its surface. The seemingly-endless silvery strip that traveled the planet's surface was unidentifiable as of yet. They see the river-like thing early on, but Campbell spots a humanoid through his telescope--this being is much like a human man, including the fact that he wore clothing. Captain Linden decides it's time for introductions, as if he senses he can trust this being, but they watch as a female and then many other people join the first man on the surface, seemingly coming out of an underground city. Linden and Campbell think their ship is out of sight, and watch a ritual that the man is performing to the setting sun. The crowd of people continues to increase, and Linden notices that the landscape is moving: trees are shifting in the ground. He and Campbell stay in the ship and observe the various types of clothing and the ritual itself, as well as the moving trees which seemed to be moving to attack the people. They are indeed warriors starting an attack, and started swinging weapons. Linden tells Campbell to start the siren on their ship to scare away the attackers, and the first man they'd seen, presumably the leader, starts towards the ship. Once they are close enough, it is obvious that the humanoids don't have eyebrows or eye lashes. Captain Linden hands the leader a medallion that plays a song, as a token of friendship. Tomboldo, the leader, starts a round of introductions through a lot of gesturing. Linden hopes to learn about the Serpent River through the people to understand its cultural significance, and these people start to ask about the siren noises. The warriors attack again and panic ensues, pushing the humans to use weapons this time. Gravgak, the guard who had been escorting the humans, is knocked down. As Linden tries to tend to him, Gravgak knocks him out with his club. Linden is unconscious for a few weeks, and Vauna, Tomboldo's daughter, spends a lot of time by the Captian's side. Linden reminds Campbell that they weren't allowed to marry anyone from this planet, but mostly in an effort to warn himself to be careful around Vauna. He learns that these people are called the Benzendellas. Tomboldo is baffled by the technology that the humans have, but Linden is not able to communicate his questions about the Serpent River. He sees Gravgak, who apologizes for the accidental injury, but from Vauna's reaction Linden is not sure if he is telling the truth. Gravgak insists on talking to Vauna in private, but Vauna's father calls them back. It is Tomboldo's thanks to the humans that gives a glimpse into the meaning of the Serpent River: he says the humans will ride with them on the rope of life, which they call Kao-Wagwattl.\vspace{0.2cm} \\  
         \hline 
         \vspace{0.2cm}
1. Captain Linden and Campbell lead first manned expedition to planet. \\
2. They observe a silver river-like feature on the planet's surface. \\
3. Campbell spots humanoid beings through his telescope. \\
4. They witness people emerging from an underground city. \\
5. Linden and Campbell observe a ritual and moving trees. \\
6. Tree-like warriors attack the humanoid people. \\
7. Linden uses ship's siren to scare away attackers. \\
8. Humans make first contact with the planet's inhabitants. \\
9. Warriors attack again, prompting humans to use weapons. \\
10. Gravgak knocks Linden unconscious. \\ 
11. Linden recovers, forms connection with Vauna. \\ 
\quad Q: How long is Linden unconscious? A: a few weeks \\
12. Humans learn about the Benzendellas and their technology.\\ 
\quad Q: What does Linden not know about? A: the Serpent River \\
13. Gravgak apologizes for injuring Linden. \\
14. Tomboldo reveals the Serpent River is called Kao-Wagwattl. \\
         \hline
    \end{tabular}
    \caption{An example reference summary from SQuALITY (top), with the Sonnet 3.5 synthetic plan and QA pairs (bottom).}
    \label{fig:full-plan-example}
\end{figure*}

\begin{figure*}[t]
    \centering
    \small
    \begin{tabular}{|p{0.95\linewidth}|}
         \hline
         \vspace{0.1cm}
Captain Linden and his lieutenant ``Split'' Campbell make up the first manned expedition from Earth to this particular planet, aiming to investigate a large silver river on its surface. [1] The seemingly-endless silvery strip that traveled the planet's surface was unidentifiable as of yet. [2] They see the river-like thing early on, but Campbell spots a humanoid through his telescope--this being is much like a human man, including the fact that he wore clothing. [3] Captain Linden decides it's time for introductions, as if he senses he can trust this being, but they watch as a female and then many other people join the first man on the surface, seemingly coming out of an underground city. [4] Linden and Campbell think their ship is out of sight, and watch a ritual that the man is performing to the setting sun. [5] The crowd of people continues to increase, and Linden notices that the landscape is moving: trees are shifting in the ground. [6] He and Campbell stay in the ship and observe the various types of clothing and the ritual itself, as well as the moving trees which seemed to be moving to attack the people. [7] They are indeed warriors starting an attack, and started swinging weapons. [8] Linden tells Campbell to start the siren on their ship to scare away the attackers, and the first man they'd seen, presumably the leader, starts towards the ship. [9] Once they are close enough, it is obvious that the humanoids don't have eyebrows or eye lashes. [10] Captain Linden hands the leader a medallion that plays a song, as a token of friendship. [11] Tomboldo, the leader, starts a round of introductions through a lot of gesturing. [12] Linden hopes to learn about the Serpent River through the people to understand its cultural significance, and these people start to ask about the siren noises. [13] The warriors attack again and panic ensues, pushing the humans to use weapons this time. [14] Gravgak, the guard who had been escorting the humans, is knocked down. [15] As Linden tries to tend to him, Gravgak knocks him out with his club. [16] Linden is unconscious for a few weeks, and Vauna, Tomboldo's daughter, spends a lot of time by the Captian's side. [17] Linden reminds Campbell that they weren't allowed to marry anyone from this planet, but mostly in an effort to warn himself to be careful around Vauna. [18] He learns that these people are called the Benzendellas. [19] Tomboldo is baffled by the technology that the humans have, but Linden is not able to communicate his questions about the Serpent River. [20] He sees Gravgak, who apologizes for the accidental injury, but from Vauna's reaction Linden is not sure if he is telling the truth. [21] Gravgak insists on talking to Vauna in private, but Vauna's father calls them back. [22] It is Tomboldo's thanks to the humans that gives a glimpse into the meaning of the Serpent River: he says the humans will ride with them on the rope of life, which they call Kao-Wagwattl. [23]\vspace{0.2cm} \\  
         \hline 
         \vspace{0.2cm}
1. Captain Linden and Campbell lead first manned expedition to planet. [1] \\ 
2. They observe a silver river-like feature on the planet's surface. [1, 2] \\
3. Campbell spots humanoid beings through his telescope. [3]\\
4. They witness people emerging from an underground city. [4]\\
5. Linden and Campbell observe a ritual and moving trees. [5, 6, 7]\\
6. Tree-like warriors attack the humanoid people. [8]\\
7. Linden uses ship's siren to scare away attackers. [9]\\
8. Humans make first contact with the planet's inhabitants. [10, 11, 12]\\
9. Warriors attack again, prompting humans to use weapons. [14]\\
10. Gravgak knocks Linden unconscious. [15, 16]\\
11. Linden recovers, forms connection with Vauna. [17, 18]\\
12. Humans learn about the Benzendellas and their technology. [19, 20]\\
13. Gravgak apologizes for injuring Linden. [21]\\
14. Tomboldo reveals the Serpent River is called Kao-Wagwattl. [23]\\
         \hline
    \end{tabular}
    \caption{An example reference summary from SQuALITY with sentence markers in brackets (top), with the Sonnet 3.5 synthetic plan with citations in brackets (bottom). This version of the coarse plans is passed to the QA pipeline described in order to extract QA pairs.}
    \label{fig:full-plan-example-w-citations}
\end{figure*}

\begin{figure*}
    \small
    \begin{tabular}{p{\linewidth}}

Human: Given a text, with enumerated sentences, devise a plan for a summary based on major events in the text. Each plan point should be a simple, short sentence (3-10 words), followed by references to all relevant sentences that can be used to validate the information it contains.

    Here are some examples:

<example>\\
Text:

Christina Aguilera had a good Valentine's Day: She announced yesterday that she's engaged to Matt Rutler, her boyfriend of three years, Radar reports. [1] Aguilera, 33, posted a photo to Facebook of her and Rutler holding hands on the beach—and her hand is sporting a huge ring. [2] "He asked and I said..." reads the caption. [3] Aguilera was previously married to Jordan Bratman, with whom she has a 6-year-old son. [4] (As they say, the couple who gets arrested together stays together.) [5]\\

Plan:\\
1. Christina Aguilera had a good Valentine's Day. [1]
2. She announced her engagement to Matt Rutler. [1]
3. She posted a photo of herself and Rutler holding hands. [2]
4. She was previously married to Jordan Bratman. [4]
5. She has a son with Bratman. [4]
6. The couple was arrested together previously. [5]\\
</example>\\

<example>\\
Text:

The US stands by the "one-China" policy, but that doesn't mean it can't sell weapons directly to Taiwan, citing the Taiwan Relations Act to ensure Taiwan can adequately defend itself—and China isn't happy about it. [1] The Obama administration announced a \$1.8 billion arms package sale to Congress on Wednesday, Reuters reports, including guided-missile frigates, anti-tank missiles, Amphibious Assault Vehicles, and \$416 million worth of guns, ammo, and other supplies. [2] The announcement came amid reports that the US had stalled the sale to avoid hearing about it from China, which still claims Taiwan as a territory, per the Wall Street Journal. [3] Reuters notes the sale comes as US-China relations simmer over the latter's man-made islands in the South China Sea and US patrols in those waters. [4] China notes it's going to sanction the companies involved in the sale (including Lockheed Martin and Raytheon), with a foreign ministry official telling Xinhua that the sale flouts international rules and "severely" damages China's sovereignty. [5] "China's government and companies will not carry out cooperation and commercial dealings with these types of companies," a ministry spokesman says. [6] A Pentagon spokesman gave the equivalent of an eyeroll Wednesday, per the New York Times, noting, "The Chinese can react to this as they see fit. [7] It's a clear-eyed, sober view of an assessment of Taiwan's defense needs. [8] There's no need for it to have any derogatory effect on our relationship with China." [9] Meanwhile, the AP notes that China has issued similar threats before, with "no evidence they've had any meaningful effect." (All this despite a lengthy handshake last month.) [10]\\
Plan:\\
1. US announced an arms package sale to Taiwan. [2]
2. China is not happy about it. [1]
3. China threatens to sanction companies involved in the sale. [5]
4. US shrugs off the threat. [9]
5. China has issued similar threats before without any meaningful effect. [10]\\
</example>\\

Assistant: Ok. How many plan points do you want me to include?

Human: This will depend on the length of the text. If the text is long you can include many plan points. Make sure each significant event or occurrence is represented in the plan.

Assistant: What are the numbers in the bracket such as [1], [2] etc at the end of each plan point?

Human: Good question. For each plan point, you are required to cite the relevant sentence number which can be used to validate the information contained in the plan point. If multiple sentences need to be cited, then separate the sentence numbers with comma such as [1, 2, 3] or [8, 10].

Assistant: Ok, so the numbers at the end of the plan point correspond to the relevant sentence numbers based on which the plan point was generated.

Human: Yes, that is correct. Please be very careful with the citation. It is very important that you get the citation correct for all of the plan points. Please note again that each sentence in the Text ends with the sentence number such as [1], [2] etc.

Assistant: Ok, I will do my best.

Human: Now it's your turn to write plans. I'll give you the text and you give me the plan with citations for each plan point. Provide your response after "Plan:".

Text:\\
\{\}

Assistant:
% Human: Given a text, devise a plan for a summary, based on major events in the text. Each plan point should be a simple, short sentence. Here are some example plans:\\
% \\

% Example 1: \\
% 1. Pope Francis visited his predecessor and exchanged Christmas greetings.\\
% 2. They chatted in Pope Benedict's retirement home.\\
% 3. They prayed together at the chapel.\\ \\

% Example 2: \\
% 1. The Costa Concordia will be refloated in the Mediterranean Sea.\\
% 2. The salvage operation is the biggest in history.\\
% 3. It will be towed to be dismantled for scrap.\\
% 4. Many passengers and crew members died in the disaster.\\
% 5. The ship will be escorted by other vessels.\\
% 6. The captain is on trial for manslaughter.\\ \\

% Example 3:\\
% 1. Khodorkovsky was released from prison on a presidential pardon.\\
% 2. He met with his family in Berlin.\\
% 3. Khodorkovsky held a news conference on Sunday.\\
% 4. He outlined his future plans.\\
% 5. He stated he will not enter into politics or seek seized oil assets.\\ \\

% Assistant: Ok. How many plan points do you want me to include? \\ \\

% Human: This will depend on the length of the text. If the text is long you can include many plan points. Make sure each significant event or occurrence is represented in the plan. \\ \\ 

% Assistant: Ok, sounds good. \\ \\

% Human: Now it's your turn to write plans. I'll give you the text and you give me the plan. Provide your response after "Plan:".\\ \\

% Text:\\
% \{summary\}\\ \\

% Assistant:\\ \\ 
% Plan:
    \end{tabular}
    \caption{Prompt supplied to Sonnet 3.5 to produce plans from summaries. For brevity, we only include two ICL examples here. Our actual prompt contains three ICL examples.}
    \label{fig:plan-prompt}
\end{figure*}

\begin{figure*}
\small
\begin{tabular}{p{0.95\linewidth}}
Human: Please summarize the following text (included within <text> and </text> tags) in up to 512 words.

Return the summary within <summary> and </summary> tags.

<text>

\{\}

</text>

Assistant:
\end{tabular}
\caption{The prompt used for summarizing documents with Sonnet 3.5.}
\label{fig:summ-prompt}
\end{figure*}

\begin{figure*}
\small
\begin{tabular}{p{0.95\linewidth}}
Baseline:\\\\
Generate a summary for the following text. Enclose the summary within <summary> and </summary> tags.\\
Text:\\
\{\}\\\\
E2E:\\\\
Generate a plan followed by a summary for the following text. Enclose the plan within <plan> and </plan> tags and enclose the summary within <summary> and </summary> tags.\\
Text:\\
\{\}\\\\
Multitask (for plans):\\\\
Generate a plan for the following text. Enclose the plan within <plan> and </plan> tags.\\
Text:\\
\{\}\\\\
Multitask (for summaries):\\\\
Generate a summary for the following text. Enclose the summary within <summary> and </summary> tags.\\
Text:\\
\{\}\\\\
\end{tabular}
\caption{The prompts we used for summarizing documents with Phi-3}
\label{fig:phi3-prompts}
\end{figure*}

\begin{figure*}
    \centering
    \small
    \begin{tabular}{p{0.95\linewidth}}
\toprule
\textbf{Instructions}

\begin{enumerate}
    \item After reading the story, write a list of ``key facts'' from the source document. A key fact should be a major narrative point in the story. Feel free to consult the associated reference summaries to refine the list.
    \item Write a list of atomic facts from the predicted summary, using the definition from \citet{kim2024fables}.
    \item Then compute each metric as follows:
\end{enumerate}

\textbf{Coverage:} For each key fact in the document, does the key fact appear in the predicted summary? Compute coverage as:
\[
\frac{\text{\# of key facts in the source document that appear in the predicted summary}}{\text{\# key facts in the source document}}
\]

\textbf{Faithfulness:} For each atomic fact in the predicted summary, is it supported in the source document? Compute faithfulness as:

\[
\frac{\text{\# of supported atomic facts in the predicted summary}}{\text{\# of atomic facts in the predicted summary}}
\]

\textbf{Conciseness:} For each atomic fact in the predicted summary, is it highly relevant to the story? More concretely, does the atomic fact appear in one of the reference summaries? Compute conciseness as:

\[
\frac{\text{\# of atomic facts in the predicted summary that appear in a reference summary}}{\text{\# of atomic facts in the predicted summary}}
\]

\textbf{Grounding:} For each plan point in the predicted plan, does the predicted summary contain / refer to it? Compute grounding as: 

\[
\frac{\text{\# of plan points appearing in the predicted summary}}{\text{\# of plan points}}
\] \\
\bottomrule
    \end{tabular}
    \caption{Evaluation rubric used for the manual evaluation task.}
    \label{fig:rubric}
\end{figure*}

\begin{figure*}
\centering
\small
\begin{tabular}{p{0.95\linewidth}}
\toprule
\begin{enumerate}[nosep]
\item Captain Linden and his lieutenant ""Split"" Campbell are exploring a planet, in particular a large, silver river.
\item They observe a group of human-like beings emerge from underground and prepare to meet them.
\item They observe moving trees, which turn out to be warriors in disguise preparing to attack the former group.
\item Linden and Campbell hit a siren on the ship, startling the attackers into retreating.
\item Linden meets the leader of the aliens, Tomboldo, and presents him a song-playing medallion.
\item Soon, they are attacked again. Linden and Campbell use their capsule bombs to dispell the warriors.
\item Their guard, Gravgak, is injured. After being awoken, Gravgak, possibly accidentally, knocks out Linden with a club.
\item Linden spends several weeks recuperating, tended by Campbell and Tomboldo's beautiful daughter Vauna.
\item While recovering, Linden reminds Campbell that marrying native inhabitants is against their mission's code of conduct.
\item Tomboldo announces to the Benzendella and the humans they will travel on the river serpent, called Kao-Wagwattl.
\end{enumerate} \\
\midrule
\begin{enumerate}[nosep]
\item Captain Linden and Splitland their ship on a strange planet.
\item They want to learn more about it.
\item They see a long silvery serpent-like object crawling on the surface.
\item Linden and Campbell are the first humans to land on the planet.
\item Linden and Campbell have been sent by EGGWE.
\item They know it is inhabited by humanoid creatures.
\item They know there is a long, cylindrical rope crawling the surface.
\item Linden orders Campbell to take a closer look at the rope.
\item Campbell reports the object is a living creature.
\item The creature is upright and wearing clothes.
\item Linden is excited to see a human-like creature on the planet.
\item Linden orders Campbell to get ready to meet the creature.
\item Linden and Campbell observe a group of natives watching the sunset.
\item The leader is wearing a red sash and headress.
\item Other members of the group are all handsome and musclar.
\item Members of the group are all wearing white fur ornaments for protection.
\item Linden and Campbell notice the trees around the natives are moving.
\item They realize the natives are not aware they are being watched.
\item A group of savage warriors suddently appear and attack the natives.
\item The natives try to defend themselves but are no match.
\item The warriors are armed with crude clubs and whips.
\item The natives rally around their leader.
\item Linden and Campbell decide to intervene to save them.
\item Linden and Campbell use a siren to scare off the warriors.
\item Linden and Campbell descend from the ship to join the natives.
\item The natives invite Linden and Campbell to their city.
\item Gravgak is a guard responsible for keeping an eye on the trees.
\item Campbell is injured when Gravgak accidentally hits him with a club.
\item Linden uses a capsule bomb to stop the attackers.
\item Gravgak is killed.
\item The natives nurse Campbell back to health.
\end{enumerate}\\
\bottomrule
\end{tabular}
\caption{(Top) Key facts extracted from a document in SQuALITY. (Bottom) Atomic facts from the same document.}
\label{fig:claim-example}
\end{figure*}

\end{document}